\documentclass{article} 
\usepackage{iclr2027_conference,times}
\iclrfinaltrue
\usepackage{amsmath,amssymb,bm}
\usepackage{booktabs}
\usepackage{array}
\usepackage{graphicx}
\usepackage{xcolor}
\usepackage{colortbl}

\usepackage{amsmath,amsfonts,bm}

\def\eqref#1{equation~\ref{#1}}

\def\1{\bm{1}}

\DeclareMathAlphabet{\mathsfit}{\encodingdefault}{\sfdefault}{m}{sl}
\SetMathAlphabet{\mathsfit}{bold}{\encodingdefault}{\sfdefault}{bx}{n}

\newcommand{\R}{\mathbb{R}}

\newcommand{\Cov}{\mathrm{Cov}}

\usepackage{hyperref}
\usepackage{url}

\newcommand{\pred}{\mathrm{pred}}
\newcommand{\pic}{\mathrm{PIC}}
\newcommand{\iso}{\mathrm{iso}}
\newcommand{\LN}{\mathrm{LN}}
\newcommand{\one}{\mathbf{1}}
\newcommand{\worldwideweb}{\raisebox{-1.3pt}{\includegraphics[height=1.05em]{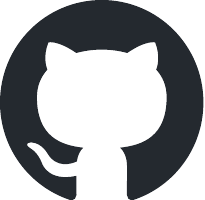}}}
\newcommand{\TBLGAP}{3.4pt}
\newcommand{\TBLSEP}{6pt}
\newcommand{\myparagraph}[1]{\par\noindent\textbf{#1}\hspace{0.6em}}

\title{Hamiltonian JEPA: Action-Conditioned World Models with an Inherited Control State}

\author{Tamim Zoabi\ \And Ameen Ali \And Lior Wolf 
 \AND \\
The Blavatnik School of Computer Science, Tel Aviv University \\
\texttt{\{tamimzoabi, ameenali023\}@mail.tau.ac.il, wolf@cs.tau.ac.il}
}
\begin{document}
\maketitle

\begin{abstract}
Planning from pixels needs more than a latent space that is stable and predictable. The state the planner scores must also be organized by how actions move the system. Joint-embedding predictive architectures (JEPAs) avoid pixel reconstruction by predicting future representations, but existing action-conditioned JEPAs ask one embedding to serve both perception and control. We introduce H-JEPA, which separates the two. A wide perceptual code is regularized toward a well-scaled isotropic geometry with a Bures-Wasserstein prior, and a fixed orthonormal slice of that code is the control state, which inherits the code's covariance without any objective of its own. The state evolves under phase-conditioned dissipative port-Hamiltonian dynamics whose input port has orthonormal columns. Port-inverse consistency (PIC) reads the executed action back through the transpose of that port. We show that this readout is exactly the rollout error projected onto the port directions, so PIC is a parameter-free reweighting of prediction error and not an auxiliary action decoder. Untying the readout from the port breaks this identity and loses half of the gain. H-JEPA matches or exceeds reconstruction-free baselines, including the action-decoding Delta-JEPA, on four pixel-based control benchmarks after at most $10$ training epochs, and its largest gain is on OGB-Cube ($91.9$ against $79.3$ percent). Ablations on PushT and OGB-Cube separate the contributions of the structured predictor, PIC, the prediction horizon, the state rank, and the anti-collapse prior.
\end{abstract}
\begin{center}
{\normalfont\fontsize{10}{13}\selectfont
  \href{https://github.com/tmz-lab/hjepa}{\worldwideweb~ Github Link}
}
\end{center}

\section{Introduction}
\label{sec:intro}

An agent that can predict the consequences of its actions can plan: it scores candidate action sequences in imagination and executes the best one \citep{ha2018world,hafner2019planet}. Trained on an offline corpus of interaction data, model-predictive control on a learned world model reaches goal images never demonstrated during training, with no reward function and no test-time gradient updates \citep{zhou2024dinowm,pldm2025}. The plan is only as good as the model's response to actions. A model that captures appearance but blurs the difference between action sequences ranks all plans alike, so action sensitivity is the central property of the learned dynamics.

Learning such models from pixels raises the question of what to predict. Reconstruction-based models regress future frames and can spend capacity on visual detail that is irrelevant to control \citep{watter2015embed,hafner2019planet,hafner2020dreamer}. Joint-embedding predictive architectures (JEPAs) instead predict future representations \citep{lecun2022path,assran2023ijepa,bardes2024vjepa}, which removes the reconstruction target but exposes training to representation collapse. LeWM \citep{lewm2026} addressed collapse for action-conditioned world models by attaching SIGReg \citep{balestriero2025lejepa}, a sketched Gaussian regularizer, to the embedding the planner scores.

This paper starts from the observation that stability is not the same as control. Matching a marginal embedding distribution certifies that no direction has died. It does not say which directions actions move, which coordinates should enter the planning metric, or how much capacity control requires. Three consequences follow. First, LeWM plans in the full embedding, so the metric that ranks candidates mixes control-relevant variation with dimensions that satisfy the prior without informing action selection. Second, the regularizer and the prediction loss act on the same coordinates, so their balance must be re-tuned whenever any other pressure changes. Third, latent-MSE objectives admit scale-degenerate solutions, since shrinking the targets makes prediction easy. Sub-JEPA \citep{subjepa2026} found the full-space prior too restrictive for low-dimensional dynamics and relaxed it to random subspaces with clear gains, but its subspaces are views for the regularizer and not an extracted state.

H-JEPA (Hamiltonian JEPA) separates the perceptual code from the control state and lets the state inherit its geometry. A wide code $h\in\R^{D}$ is placed on a sphere by per-sample affine-free normalization and regularized toward isotropy with a Bures-Wasserstein moment penalty \citep{gelbrich1990}. A fixed orthonormal projection $U$ defines the state $s=U^{\top}h$, whose covariance is a fixed function of the code's covariance, so no separate state-space objective is needed. On this state we place a dissipative port-Hamiltonian step whose input port has orthonormal columns. \emph{Port-inverse consistency} (PIC) reads the executed action back from the realized displacement through the transpose of the port, which is its exact pseudoinverse. The readout has no parameters and no free scale, and its residual equals the rollout error projected onto the port directions, so PIC reweights prediction error toward the directions in which actions act.

H-JEPA matches or exceeds reconstruction, full-space-prior, and action-decoding baselines on Two-Room, Reacher, PushT, and OGB-Cube under the same planning protocol, with the largest gain on OGB-Cube ($91.9$ percent against $79.3$ for the strongest baseline). Our contributions are as follows. (i) A two-space world model in which one distributional objective shapes perception and a fixed orthonormal map carries covariance to the planner-facing state, with an exact description of the inherited covariance (\S\ref{sec:perception} to \S\ref{sec:iso}). (ii) An analysis of anti-collapse penalties near vanishing variance, showing that the Bures-Wasserstein penalty keeps a constant restoring force on each standard deviation while SIGReg and the Euclidean covariance penalty lose theirs (\S\ref{sec:iso}). (iii) A phase-conditioned dissipative port-Hamiltonian predictor with an orthonormal input port, and PIC, which we show to be an exact error reweighting, with a control showing that the tie between readout and port accounts for half of its gain (\S\ref{sec:dyn}, \S\ref{sec:obj}, \S\ref{sec:abl}). (iv) Planning, probing, and ablation results at a matched representation on two tasks, with diagnostics showing that the planner ranks candidates along directions beyond the probed physical variables (\S\ref{sec:exp}).

\section{Related work}
\label{sec:related}

\myparagraph{JEPA world models.}
\citet{lecun2022path} proposed prediction in representation space as the basis of world models, validated by I-JEPA and V-JEPA \citep{assran2023ijepa,bardes2024vjepa}. LeJEPA \citep{balestriero2025lejepa} motivates isotropic Gaussian embeddings and enforces them with SIGReg. LeWM \citep{lewm2026} brought this program to action-conditioned world models from pixels, and Sub-JEPA \citep{subjepa2026} improved planning by applying the Gaussian constraint in $K$ random subspaces, with $K$ selected per benchmark. Delta-JEPA \citep{deltajepa2026} adds action sensitivity through a learned linear decoder on latent differences, and sensorimotor world models \citep{smwm2026} make inverse dynamics the perception objective. H-JEPA differs in three respects. The subspace is a single fixed slice that \emph{is} the control state, the prior stays full-space at LeWM's placement and reaches the state by inheritance, and the inverse readout is the forward model's own input port.

\myparagraph{Latent dynamics models for control.}
E2C, DVBF, World Models, PlaNet, and Dreamer learn latent dynamics through pixel reconstruction and often reward prediction \citep{watter2015embed,karl2017dvbf,ha2018world,hafner2019planet,hafner2020dreamer}. DINO-WM plans with a frozen pretrained encoder \citep{zhou2024dinowm}, and PLDM trains an end-to-end pixel JEPA with a VICReg-style objective \citep{pldm2025}. H-JEPA uses neither pixels nor rewards as targets and trains its encoder end-to-end without teachers or stop-gradients \citep{grill2020byol}. Its collapse defense is a moment-matching penalty in the Barlow Twins and VICReg family \citep{zbontar2021barlow,bardes2022vicreg}, with the Euclidean covariance metric replaced by Bures-Wasserstein on Gaussians \citep{bhatia2019bures}.

\myparagraph{Structured dynamics.}
Hamiltonian neural networks learn a scalar energy whose gradient generates conservative flow \citep{greydanus2019hnn}, with controlled and generalized variants \citep{zhong2020symplectic,finzi2020weak,chen2021nsf} and image-based generative forms \citep{toth2020hgn}. That literature targets a specified physical state or reconstruction. H-JEPA instead adopts the port-Hamiltonian form \citep{vanderschaft2014ph} as an inductive bias for a reconstruction-free representation trained for planning, and uses port orthonormality to remove a scale freedom that would otherwise interact with the representation objective.

\myparagraph{Inverse dynamics and state discovery.}
Recovering actions from transitions is a classical route to control-relevant features \citep{agrawal2016learning,pathak2017curiosity}, as are DeepMDP and bisimulation objectives \citep{gelada2019deepmdp,zhang2021bisim}. In exogenous-block MDPs, forward models with multi-step inverse prediction identify a control-endogenous state under stated assumptions \citep{lamb2022acstate}, with known limits on when multi-step inverse prediction is sufficient \citep{levine2024acdf}, and displacement-level action decoding appears in recent JEPA world models \citep{deltajepa2026,smwm2026}. PIC ties the readout to the transpose of the forward input map. Tying a readout to the transpose of an input map is standard practice for language-model embeddings \citep{press2017tying}, where the transpose is not the inverse of that map in general \citep{pit2026}. In H-JEPA the port is constrained to be orthonormal, which makes the transpose its exact pseudoinverse. The bottleneck view relates to the information bottleneck \citep{tishby1999information,alemi2017deep}, predictive state representations \citep{littman2001psr}, and balanced truncation \citep{moore1981balanced}.
\section{Method}
\label{sec:method}
\begin{table}[t]
\centering
\caption{Design choices we relaxed during development, the failure each one produced in this architecture, and the constraint adopted instead. These observations motivate the constraints of \S\ref{sec:method} and complement the controlled ablations of \S\ref{sec:abl}.}
\label{tab:negatives}
\footnotesize
\setlength{\tabcolsep}{4pt}
\setlength{\aboverulesep}{0pt}
\setlength{\belowrulesep}{0pt}
\renewcommand{\arraystretch}{1.25}
\begin{tabular}{@{}>{\raggedright\arraybackslash}p{3.3cm}>{\raggedright\arraybackslash}p{5.9cm}>{\raggedright\arraybackslash}p{4.0cm}@{}}
\toprule
Relaxed choice & Observed failure & Adopted constraint \\
\midrule
\rowcolor{gray!12}
BatchNorm in the projector, or a standardizer on $s$ & Hides collapse: $s$ keeps unit marginals while $h$ degenerates to rank one & Per-sample affine-free $\LN_0$, no normalizer on $s$ (\S\ref{sec:perception}, \S\ref{sec:bottleneck}) \\
LSUV-style whitening of the projector activations at initialization \citep{mishkin2016lsuv} & Amplifies projector weights about $10^4$-fold, point-mass collapse within tens of steps & Isotropy as a loss with bounded per-step influence (\S\ref{sec:iso}) \\
\rowcolor{gray!12}
Learnable projection $U$ & Co-adapts with the losses and collapses the rank of the state & Fixed seeded orthonormal $U$ (\S\ref{sec:bottleneck}) \\
Learned action embedding & The PIC regression target collapses to effective rank about $1.3$ & Raw normalized actions (\S\ref{sec:perception}) \\
\rowcolor{gray!12}
SIGReg, or the Euclidean penalty $\|\widehat{\Sigma}-I\|_F^2$ & The restoring force fades as variance shrinks, collapse within about $50$ to $100$ steps & Bures-Wasserstein prior (\S\ref{sec:iso}) \\
Unconstrained port gain & The gain trades off against the encoder scale, so optimization moves scale and not content & Orthonormal port via differentiable QR (\S\ref{sec:dyn}) \\
\bottomrule
\end{tabular}
\end{table}

\myparagraph{Setting.}
We are given offline trajectories of RGB frames $x_t$ and continuous actions, with no rewards and no simulator state. The goal is a latent model in which a planner can roll out candidate action sequences and rank them by distance to an encoded goal image. An encoder produces a \emph{perceptual code} $h_t$ (\S\ref{sec:perception}). A fixed linear slice of the code is the \emph{control state} $s_t$, the only quantity the planner scores (\S\ref{sec:bottleneck}). One distributional prior acts on the code and keeps it from collapsing (\S\ref{sec:iso}). A port-Hamiltonian predictor advances the state under actions (\S\ref{sec:dyn}). Training adds port-inverse consistency (PIC), which asks the predictor's own action-input map to recover the executed action from the observed displacement (\S\ref{sec:obj}). One model step is
\begin{equation}
h_t = E(x_t),\qquad
s_t = U^{\top}h_t,\qquad
v_t = s_t - s_{t-1},\qquad
c_t = [s_t,v_t],\qquad
\hat s_{t+1} = \mathcal{F}_\theta(s_t,\bar a_t \mid c_t).
\label{eq:loop}
\end{equation}
Reading \eqref{eq:loop} from left to right, $E$ is the image encoder and $h_t\in\R^{D}$ is the code of the single frame $x_t$. The matrix $U\in\R^{D\times r}$ has orthonormal columns and is never trained, and $s_t\in\R^{r}$ with $r\le D$ is the control state. Since $s_t$ is computed from one frame it carries no motion information, so we form the finite-difference velocity $v_t\in\R^{r}$ and concatenate it with the state into the \emph{phase} $c_t\in\R^{2r}$ (square brackets denote concatenation). The action $\bar a_t\in\R^{d_a}$ is the normalized control applied between frames $t$ and $t+1$. The predictor $\mathcal{F}_\theta$ with parameters $\theta$ advances $s_t$ by one step under $\bar a_t$. The notation $\mathcal{F}_\theta(\cdot\mid c_t)$ means that the vector fields inside the predictor are functions of the phase, while only the state is advanced. The velocity is recomputed from consecutive states, with $v_0=0$ at the start of a window, and in open-loop rollouts it is formed from predictions, $\hat v_{t+i}=\hat s_{t+i}-\hat s_{t+i-1}$. An overview figure is given in Appendix~\ref{app:arch}.

One rule organizes the design. The distributional objective acts only on perception, every map between the code and the planner is fixed or explicitly constrained, and no batch statistic or learned gain sits on that path. Each constraint answers a failure we observed when it was relaxed (Table~\ref{tab:negatives}).

\subsection{Perception on a sphere}
\label{sec:perception}

Following the LeWM perception stack, each frame is encoded independently as
$$
h_t=\LN_0\!\bigl(\mathrm{MLP}(\mathrm{ViT}(x_t))\bigr)\in\R^D ,
\qquad
\LN_0(z)=\sqrt{D}\,\frac{z-\operatorname{mean}(z)\one}{\|z-\operatorname{mean}(z)\one\|_2},
$$
with a ViT-T/14 trained from scratch \citep{dosovitskiy2021vit} and an MLP projector at $D=192$. $\LN_0$ is LayerNorm without learned scale or bias. It centers each sample across its coordinates and fixes $\|h\|_2=\sqrt{D}$, so every code satisfies $e^{\top}h=0$ with $e=\one/\sqrt{D}$ and lies on a common sphere. Fixing the radius removes a scale freedom that would let the encoder inflate or shrink transition magnitudes instead of improving geometry, so the prior of \S\ref{sec:iso} redistributes variance and cannot grow it.

Each model step spans $m$ environment steps (the frameskip), and the action is the mean commanded control over that window, $\bar a_t=\tfrac{1}{m}\sum_{j=1}^{m}a_{t,j}$, in normalized coordinates. We use no learned action embedding, which co-adapted with the dynamics in our experiments (Table~\ref{tab:negatives}).

\subsection{A control state with inherited covariance}
\label{sec:bottleneck}

The control state is $s_t=U^{\top}h_t\in\mathbb{R}^r$, where $U$ is drawn once from a seeded QR factorization of a Gaussian matrix. Because $U$ is fixed and linear,
\begin{equation}
\Cov(s)=U^{\top}\Cov(h)\,U ,
\label{eq:inherit}
\end{equation}
so the covariance geometry imposed on the perceptual code is inherited by the control state without introducing a separate state-space objective or normalization. We refer to this property as \emph{covariance inheritance}. At the prior's target $\Cov(h)=I-ee^{\top}$ (\S\ref{sec:iso}), Eq.~\eqref{eq:inherit} gives

$$
\Cov(s)=I_r-(U^{\top}e)(U^{\top}e)^{\top}.
$$

Thus the target covariance is isotropic up to a single exceptional direction, whose eigenvalue is $1-|U^{\top}e|_2^2$. For a random orthonormal $U$, $|U^{\top}e|_2^2$ concentrates around $r/D$, so the deviation from isotropy is controlled directly by the relative state dimension. When $r<D$, the state is therefore nearly isotropic with one direction receiving the corresponding shrinkage. When $r=D$, the centering direction $e$ is contained in the span of $U$, giving one exact null direction. Trained codes need not reach the prior exactly, and Appendix~\ref{app:geometry} reports the resulting covariance spectra and the predicted null direction. Keeping $U$ fixed also prevents the projection from co-adapting with the dynamics, while experiments with a learnable $U$ instead produced degenerate low-rank states (Table~\ref{tab:negatives}).

\subsection{A Bures-Wasserstein isotropy prior}
\label{sec:iso}

Let $\mathcal{H}=\{h_j\}_{j=1}^{n}$ be the codes of all frames in a batch, pooled over batch and time, with empirical mean $\hat\mu=\tfrac1n\sum_j h_j$ and covariance $\widehat{\Sigma}=\tfrac1n\sum_j (h_j-\hat\mu)(h_j-\hat\mu)^{\top}$. Since every code satisfies $e^{\top}h=0$, the covariance always has a zero eigenvalue along $e$, and the most isotropic covariance available is the projector $P=I-ee^{\top}$ onto the hyperplane orthogonal to $e$. We penalize the squared $2$-Wasserstein distance $W_2^2$ between the Gaussian with the empirical moments and the Gaussian with the target moments, which has the closed Bures-Wasserstein form \citep{gelbrich1990,bhatia2019bures}:
\begin{equation}
\boxed{\ 
\mathcal{L}_{\iso}(\mathcal{H})
= \frac{1}{D}\,W_2^2\bigl(\mathcal{N}(\hat\mu,\widehat{\Sigma}),\,\mathcal{N}(0,P)\bigr)
= \frac{1}{D}\Bigl(\|\hat\mu\|_2^{2}
+\textstyle\sum_{i=1}^{D-1}\bigl(\sqrt{\hat\lambda_i}-1\bigr)^2\Bigr)
\ }
\label{eq:iso}
\end{equation}
where $\hat\lambda_1,\dots,\hat\lambda_{D-1}\ge 0$ are the eigenvalues of $\widehat{\Sigma}$ on that hyperplane. The second equality holds because $\widehat{\Sigma}$ and $P$ commute. Equation~\ref{eq:iso} constrains only the first two moments and does not require $h$ to be Gaussian. It is the model's only distributional objective.

\myparagraph{Why Bures-Wasserstein.}
Write $\sigma_i=\sqrt{\lambda_i}$ for the standard deviation of the code along one covariance direction. In H-JEPA the $D-r$ code directions outside the slice receive no gradient from prediction or PIC, so the prior alone must hold them open, and what matters is the force a penalty exerts on $\sigma_i$ as $\sigma_i\to0$. The three candidates behave differently:
\begin{equation}
\underbrace{\tfrac{\partial}{\partial\sigma}(\sigma-1)^2=2(\sigma-1)\to-2}_{\text{Bures-Wasserstein}},
\quad
\underbrace{\tfrac{\partial}{\partial\sigma}(\sigma^2-1)^2=4\sigma(\sigma^2-1)\to0}_{\text{Euclidean }\|\widehat{\Sigma}-I\|_F^2},
\quad
\underbrace{\tfrac{\partial}{\partial\sigma}T(\sigma)=O(\sigma)\to0}_{\text{SIGReg}}.
\label{eq:forces}
\end{equation}
Bures-Wasserstein is a quadratic penalty on standard deviations, close in spirit to the variance term of VICReg \citep{bardes2022vicreg}, and keeps a constant pull on a dying direction. The Euclidean penalty is quadratic in the variance, so its pull fades linearly with $\sigma$. SIGReg, the sketched Epps-Pulley normality statistic $T$ \citep{epps1983test} used by LeJEPA and LeWM, is a smooth bounded function of the samples and is therefore also quadratic in $\sigma$ near a point mass (Appendix~\ref{app:collapse}). This matches what we saw in training. In collapse runs within H-JEPA the SIGReg score saturated as the code concentrated and its gradient became negligible, exactly where latent prediction admits a constant solution, and Table~\ref{tab:mechanism} shows the same contrast in planning success.

\subsection{Phase-conditioned port-Hamiltonian dynamics}
\label{sec:dyn}

Instead of an unconstrained transition map, the predictor has the form of a port-Hamiltonian system \citep{vanderschaft2014ph}: autonomous motion is generated by a scalar energy and damped by a dissipation term, and the action enters through a separate input map called the \emph{port}. We use this form as an inductive bias and do not assume that the latent coordinates are physical positions and momenta. For even $r$, the predictor consists of
\begin{equation}
J =
\begin{bmatrix}
0 & -I_{r/2} \\
I_{r/2} & 0
\end{bmatrix},
\qquad
H_\theta : \R^{2r} \rightarrow \R,
\qquad
R_\theta(c) = L_\theta(c) L_\theta(c)^\top \succeq 0,
\qquad
G_\theta(c) \in \R^{r \times d_a}.
\label{eq:parts}
\end{equation}
Here $J\in\R^{r\times r}$ is the constant canonical symplectic matrix, with $I_{r/2}$ the identity of size $r/2$. It is skew-symmetric, so the flow $J\nabla_s H$ conserves $H$ in continuous time. $H_\theta$ is a scalar energy computed by an MLP from the phase $c=[s,v]$, and $\nabla_s H_\theta$ denotes its gradient with respect to the state argument only, with $v$ treated as a constant input. $L_\theta(c)\in\R^{r\times r}$ is a lower-triangular matrix output by a second MLP, so the dissipation matrix $R_\theta(c)$ is positive semidefinite by construction and can only remove energy. $G_\theta(c)$ is the input port, which maps an action in $\R^{d_a}$ to a state displacement. It is the output of a third MLP followed by a differentiable thin QR factorization, so its columns are orthonormal, $G_\theta(c)^{\top}G_\theta(c)=I_{d_a}$ (it lies on the Stiefel manifold). All three networks take the phase as input, which is how velocity information reaches the dynamics. The continuous-time field $\dot s=(J-R_\theta(c))\nabla_s H_\theta(s,v)+G_\theta(c)\bar a$ is discretized with one explicit Euler step of size $\Delta t=1$:
\begin{equation}
\hat s_{t+1}
=
\mathcal{F}_\theta\bigl( s_t, \bar a_t \mid c_t \bigr)
=
s_t
+
\Delta t
\Big[
\underbrace{\bigl( J - R_\theta(c_t) \bigr) \nabla_s H_\theta(s_t, v_t)}_{\text{passive flow}}
+
\underbrace{G_\theta(c_t)\, \bar a_t}_{\text{control flow}}
\Big].
\label{eq:step}
\end{equation}
Continuous-time energy properties need not hold exactly after Euler discretization (\S\ref{sec:limits}).

\myparagraph{The orthonormal port fixes action geometry and scale.}
An unconstrained input map $B(c)$ can be written as $G(c)S(c)$ with orthonormal $G$ and gain $S$. The gain can trade off against latent scale and PIC readout, causing scale drift without improving content (Table~\ref{tab:negatives}). Constraining the port to orthonormal columns removes this freedom. The action Jacobian $\partial \mathcal{F}/\partial \bar a=\Delta t\,G_\theta(c)$ then has all singular values equal to $\Delta t$, so one unit of normalized action produces one unit of direct latent displacement. State-dependent control effectiveness is therefore expressed through the port orientation and state geometry \citep{murray1994mathematical}.

\subsection{Training objective}
\label{sec:obj}

Training uses three terms,
\begin{equation}
\boxed{
\mathcal{L}
=
\mathcal{L}_{\pred}
+
\mathcal{L}_{\pic}
+
\mathcal{L}_{\iso}(\mathcal{H})
} ,
\label{eq:loss}
\end{equation}
with no pixel reconstruction, learned inverse model, contrastive loss, teacher network, or normalization of $s$.

\myparagraph{Open-loop prediction.}
We supervise the same multi-step composition the planner uses. Let $\mathcal{T}$ be the set of rollout start times in a batch. From an encoded state $s_t$, the model rolls forward for $K_p$ steps under the executed actions, feeding back its own predictions, and matches the encoded future at every step:
\begin{equation}
\mathcal{L}_{\pred}
=
\frac{1}{|\mathcal{T}| K_p}
\sum_{t \in \mathcal{T}}
\sum_{i=1}^{K_p}
\tfrac{1}{r}
\left\|
\delta_{t+i}
\right\|_2^2 ,
\quad
\delta_{t+i}=s_{t+i}-\hat s_{t+i},
\quad
\hat s_{t+i}
=
\mathcal{F}_\theta
\bigl(
\hat s_{t+i-1},
\bar a_{t+i-1}
\mid
\hat c_{t+i-1}
\bigr),
\label{eq:pred}
\end{equation}
with $\hat s_t=s_t$ and $\hat c_{t+i}=[\hat s_{t+i},\hat v_{t+i}]$. Squared errors are averaged over coordinates, so all three terms are of order one. For $K_p=1$ this is the next-latent regression of LeWM. We use $K_p=5$, equal to the planning horizon, so training penalizes the composed rollouts that the planner ranks.

\myparagraph{Port-inverse consistency (PIC).}
The port maps an action into a state displacement. PIC asks the same port, transposed, to map the observed displacement back to the action. At rollout step $i$, we subtract the passive flow from the observed displacement and read the remainder through the transpose of the port:
\begin{equation}
\hat u_{t+i}
=
G_\theta(\hat c_{t+i})^{\top}
\left[
\frac{s_{t+i+1} - \hat s_{t+i}}{\Delta t}
-
\bigl( J - R_\theta(\hat c_{t+i}) \bigr)
\nabla_s H_\theta \bigl( \hat s_{t+i}, \hat v_{t+i} \bigr)
\right],
\label{eq:pic-decode}
\end{equation}
and penalize its distance to the executed action over $K_{\pic}$ rollout steps:
\begin{equation}
\boxed{
\mathcal{L}_{\pic}
=
\frac{1}{|\mathcal{T}|K_{\pic}}
\sum_{t\in\mathcal{T}}
\sum_{i=0}^{K_{\pic}-1}
\tfrac{1}{d_a}
\left\|
\hat u_{t+i} - \bar a_{t+i}
\right\|_2^2
} .
\label{eq:pic}
\end{equation}
Because the port has orthonormal columns, $G_\theta^{\top}$ is its exact pseudoinverse, as in tied input-output maps \citep{press2017tying}, so the readout needs no learned inverse model and has no scale of its own. Substituting the Euler step \eqref{eq:step} into \eqref{eq:pic-decode} (Appendix~\ref{app:pic}) gives
\begin{equation}
\hat u_{t+i} - \bar a_{t+i}
=
\tfrac{1}{\Delta t}\,
G_\theta(\hat c_{t+i})^{\top}\delta_{t+i+1},
\label{eq:pic-residual}
\end{equation}
where $\delta_{t+i+1}$ is the rollout error of \eqref{eq:pred}. The prediction and PIC terms therefore combine into a single quadratic form in that error,
\begin{equation}
\mathcal{L}_{\pred}+\mathcal{L}_{\pic}
=
\frac{1}{|\mathcal{T}|K_p\,r}\sum_{t\in\mathcal{T}}\sum_{i=1}^{K_p}
\delta_{t+i}^{\top}\Bigl(I+\tfrac{r}{d_a\,\Delta t^{2}}\,G_{t+i-1}G_{t+i-1}^{\top}\Bigr)\delta_{t+i} ,
\label{eq:pic-metric}
\end{equation}
when $K_{\pic}=K_p$, with $G_{t+i-1}=G_\theta(\hat c_{t+i-1})$.

Equation~\ref{eq:pic-metric} tells us what PIC actually measures. PIC is not a separate task of predicting actions. It is the same rollout error as $\mathcal{L}_{\pred}$, restricted to the $d_a$ port directions through which the action enters the dynamics. Adding the two losses therefore gives a single rollout error, measured with a state-dependent metric that counts errors along the port directions $1+r/(d_a\Delta t^2)$ times more than errors along any other direction.

This view also explains why the model cannot satisfy PIC through a shortcut. There are two obvious shortcuts, and neither works. The first is to rescale the readout until actions become easy to decode. This fails because actions are in fixed normalized units, the port has unit singular values, and the scale of $s$ is inherited rather than set by the readout. The second is to rotate the port away from the directions where the rollout is wrong. This fails because the same port also injects the action in Equation~\eqref{eq:step}, so a port that no longer points at the error makes the prediction worse and raises $\mathcal{L}_{\pred}$.

Both arguments rest on one design choice: the action is read out through the same port $G_\theta$ that injected it. \S\ref{sec:abl} tests this directly by replacing $G_\theta^{\top}$ in the readout with a fixed random orthonormal matrix, which breaks the link between injection and readout. We use $K_{\pic}=K_p=5$ throughout.

\myparagraph{Planning.}
At test time we plan with the cross-entropy method (CEM) in a model-predictive control loop \citep{rubinstein1999cem}. The goal image is encoded to $s_{\mathrm{goal}}=U^{\top}E(x_{\mathrm{goal}})$, candidate action sequences are rolled out with \eqref{eq:step} for $K_p$ steps, and each is scored by the terminal distance $\|\hat s_{t+K_p}-s_{\mathrm{goal}}\|_2^2$. The phase conditions the dynamics but is not part of the cost. Main results follow the Delta-JEPA budget of $500$ episodes. Controlled ablations use the lower-cost LeWM budget with otherwise identical planning settings.

\section{Experiments}
\label{sec:exp}

\myparagraph{Setup.}\label{sec:setup}
We evaluate on the four pixel-based planning benchmarks of the LeWM suite: Two-Room navigation \citep{lewm2026}, DeepMind Control Reacher \citep{tassa2018dmc}, PushT \citep{chi2023diffusion}, and OGBench Cube manipulation \citep{park2025ogbench}, hereafter OGB-Cube. All models train on the same offline image-action trajectories. We compare against PLDM \citep{pldm2025}, LeWM \citep{lewm2026}, Sub-JEPA \citep{subjepa2026}, and Delta-JEPA \citep{deltajepa2026} under the Delta-JEPA evaluation protocol of $500$ test episodes per task \citep{deltajepa2026}, in which Delta-JEPA is trained from scratch for $50$ epochs. We train H-JEPA for $10$ epochs on every task. One held-out pilot seed, excluded from the reported averages, is used only to choose which epoch's checkpoint to evaluate on each task, and this choice is then fixed for the remaining seeds: epoch $8$ on Two-Room, $6$ on Reacher, $10$ on PushT, and $3$ on OGB-Cube. H-JEPA uses AdamW \citep{loshchilov2019adamw}, windows of $T=2+K_p=7$ frames, and $K_p=K_{\pic}=5$. The only task-specific architectural choice is the state rank: $32$ on Two-Room and OGB-Cube, $64$ on Reacher, and $192$ on PushT, chosen once from the lower-budget sweep on a held-out seed in \S\ref{sec:abl} and then fixed for the $500$-episode evaluation. Per-benchmark capacity selection is common in this suite: Sub-JEPA selects its number of subspaces per task ($K=32$, except $K=16$ on PushT, where $K=32$ fails) \citep{subjepa2026}. All other hyperparameters are shared (Appendix~\ref{app:config}).

\begin{table}[t]
\centering
\caption{Planning success rate (\%) on the four LeWM benchmarks over $500$ test episodes per task, mean $\pm$ std over three training seeds. Baselines are evaluated under the Delta-JEPA protocol, in which Delta-JEPA trains for $50$ epochs. H-JEPA trains for $10$ epochs and is evaluated at the per-task checkpoint epoch selected on a held-out seed (\S\ref{sec:setup}).}
\label{tab:main}
\small
\begin{tabular}{@{}lcccc@{}}
\toprule
Method & Two-Room & Reacher & PushT & OGB-Cube \\
\midrule
PLDM & $93.73\pm 1.03$ & $64.33\pm 2.14$ & $76.13\pm 1.70$ & $57.27\pm 1.53$ \\
LeWM & $74.93\pm 0.42$ & $79.87\pm 0.90$ & $84.53\pm 1.50$ & $64.13\pm 1.89$ \\
Sub-JEPA & $90.60\pm 0.53$ & $81.00\pm 2.40$ & $63.73\pm 0.12$ & $62.67\pm 1.45$ \\
Delta-JEPA & $\mathbf{100.00\pm 0.00}$ & $81.33\pm 0.50$ & $89.07\pm 1.90$ & $79.27\pm 1.81$ \\
H-JEPA (ours) & $\mathbf{100.00\pm 0.00}$ & $\mathbf{86.13\pm 0.23}$ & $\mathbf{90.40 \pm 0.60}$ & $\mathbf{91.93\pm 1.30}$ \\
\bottomrule
\end{tabular}
\end{table}

\myparagraph{Planning results.}\label{sec:results}
Table~\ref{tab:main} reports planning success. H-JEPA is best or tied on every task. It saturates Two-Room together with Delta-JEPA, is within one standard deviation of Delta-JEPA on PushT, and leads by $4.8$ points on Reacher and $12.7$ points on OGB-Cube. H-JEPA trains for $10$ epochs, against $50$ for Delta-JEPA. The Sub-JEPA paper reports higher numbers for itself ($95.0$, $84.0$, $89.0$, and $76.3$ on the four tasks) under a different protocol with $50$ evaluation episodes, against $500$ here \citep{subjepa2026}. H-JEPA is above those numbers as well. As a check on identical episodes, a paired comparison against the official LeWM checkpoint on $N{=}200$ matched episodes gives $88.5$ against $82.0$ on PushT (McNemar $p\approx 0.053$). Absolute rates in this paired test are not comparable to Table~\ref{tab:main}. Because Table~\ref{tab:main} changes representation and predictor together, \S\ref{sec:abl} separates them.

\myparagraph{What the state encodes.}\label{sec:probe}
Physical probing (Appendix~\ref{app:probe}) shows a consistent division of labor. H-JEPA is strongest on the directly controlled variable (PushT agent location, linear MSE $0.028$ against $0.062$ and $0.065$), while LeWM and Sub-JEPA decode the PushT block pose better. On OGB-Cube, H-JEPA is far stronger on joint velocity and end-effector position, and the methods are at parity on Reacher. Planning quality is thus not monotone in object-pose decodability.

\myparagraph{What the predictor learns.}
Appendix~\ref{app:ph_audit} audits the learned fields on held-out trajectories. The balance between passive and control flow in \eqref{eq:step} is task dependent. On PushT the control flow is only $0.13$ times the passive flow, the passive flow is dominated by dissipation, and linear probes from the passive flow, control flow, and energy gradient each explain $88$ to $90$ percent of the physical state increment. On OGB-Cube the control flow is $1.53$ times the passive flow. On Reacher and OGB-Cube the component-wise physical alignment is weaker, even though task variables remain strongly encoded in $s$.

\subsection{Ablations and analysis}
\label{sec:abl}

\myparagraph{Predictor class at a matched representation.}
The JEPA baselines use a causal AdaLN transformer over three latents. We keep the H-JEPA perception stack, the fixed $U$, the state $s$, and $K_p{=}5$, and swap only the dynamics for that transformer (Table~\ref{tab:predictor}). With $K_p{=}5$, LeWM's transformer on its native SIGReg embedding reaches $20$ percent. The same transformer on the H-JEPA representation reaches $84$ percent. Replacing it by the port-Hamiltonian predictor with PIC raises success to $92$ percent at $r{=}64$ while shrinking the dynamics module from $3.3$M to $0.86$M parameters. The selected $r{=}192$ model reaches $94$ percent with $11.6$M parameters in total, below the $18.0$M of LeWM. The last swap changes temporal-state construction and action injection together with the predictor, so it compares complete predictor formulations at a matched representation and horizon.

\begin{table}[t]
\centering
\caption{Predictor class on PushT at a matched representation and horizon ($K_p{=}5$ in all rows). Parameters are total model followed by dynamics module. H-JEPA rows share the ViT-T/14 encoder and $0.79$M projector.}
\label{tab:predictor}
\small
\begin{tabular}{@{}lccc@{}}
\toprule
Predictor & Params total / dyn. & Success & Representation \\
\midrule
LeWM, native embedding & $18.0$M / $10.8$M & $20\%$ & SIGReg \\
AdaLN transformer on H-JEPA, $r{=}64$ & $9.6$M / $3.3$M & $84\%$ & H-JEPA \\
Port-Hamiltonian $+$ PIC, $r{=}64$ & $7.2$M / $0.86$M & $92\%$ & H-JEPA \\
Port-Hamiltonian $+$ PIC, $r{=}192$ & $11.6$M / $5.4$M & $\mathbf{94\%}$ & H-JEPA \\
\bottomrule
\end{tabular}
\end{table}

\begin{table}[t]
\small
\setlength{\tabcolsep}{0pt}
\begin{minipage}[t]{0.47\linewidth}
\caption{Mechanism ablations, success (\%) at $N{=}300$. H is the port-Hamiltonian predictor. $^\dagger$Value from Table~\ref{tab:main}.}
\label{tab:mechanism}
\begin{tabular*}{\linewidth}{@{}l@{\extracolsep{\fill}}cc@{}}
\toprule
Configuration & PushT & Cube \\
\midrule
H $+$ PIC, prior on $h$ (full) & $\mathbf{93.3}$ & $\mathbf{91.9}\rlap{$^\dagger$}$ \\
\addlinespace[\TBLGAP]
MLP $+$ PIC & $85.0$ & $84.3$ \\
H, no PIC & $75.0$ & $89.0$ \\
MLP, no PIC & $32.0$ & \\
H $+$ PIC, untied readout $Q$ & $83.7$ & \\
\addlinespace[\TBLGAP]
H $+$ PIC, prior on $s$ & $86.3$ & \\
H $+$ PIC, SIGReg prior & $36.7$ & \\
\bottomrule
\end{tabular*}
\end{minipage}\hfill
\begin{minipage}[t]{0.49\linewidth}
\caption{Horizons on PushT, success (\%).}
\label{tab:horizon}
\begin{tabular*}{\linewidth}{@{}l@{\extracolsep{\fill}}ccccc@{}}
\toprule
$(K_p,K_{\pic})$ & $(1,0)$ & $(1,1)$ & $(3,3)$ & $(5,1)$ & $(5,5)$ \\
\midrule
Success & $12$ & $34$ & $\mathbf{94}$ & $90$ & $\mathbf{94}$ \\
\bottomrule
\end{tabular*}
\par\vspace{\TBLSEP}
\caption{State rank sweep, success (\%).}
\label{tab:rank}
\begin{tabular*}{\linewidth}{@{}l@{\extracolsep{\fill}}cccc@{}}
\toprule
Task & $r{=}16$ & $r{=}32$ & $r{=}64$ & $r{=}192$ \\
\midrule
Two-Room & & $\mathbf{100}$ & $100$ & $100$ \\
Reacher & $72$ ($80$) & $82$ ($84$) & $\mathbf{94}$ & $78$ \\
PushT & & $84$ ($81$) & $92$ & $\mathbf{94}$ \\
OGB-Cube & & $\mathbf{96}$ & & $76$, $78$ \\
\bottomrule
\end{tabular*}
\end{minipage}
\end{table}

\myparagraph{Structured dynamics and PIC.}
Table~\ref{tab:mechanism} isolates the remaining choices at matched rank on PushT ($r{=}64$, epoch $10$) and OGB-Cube ($r{=}32$, seed $11$), both at $N{=}300$. The structured predictor helps by a similar amount on both: with PIC in place, replacing the port-Hamiltonian step by an unstructured MLP costs $8.3$ points on PushT ($93.3$ to $85.0$) and $7.6$ on OGB-Cube ($91.9$ to $84.3$). The contribution of PIC is task dependent. Removing it costs $18.3$ points on PushT and $2.9$ on OGB-Cube, where the full-model reference is the main-benchmark value and the gap is within evaluation noise at $N{=}300$. This agrees with the reading of PIC as a reweighting \eqref{eq:pic-metric} and with the predictor audit. On PushT the control flow is a small part of each displacement ($0.13$ of the passive flow), so an unweighted rollout loss is dominated by passive error and says little about the action-coupled part, and extra weight on the port directions matters. On OGB-Cube the control flow is the larger part ($1.53$), so the rollout loss already emphasizes it, and the extra weight from PIC is also smaller there ($7.4$ against $33$, since $r=32$ and $d_a=5$). Without the Hamiltonian structure, PIC matters most: the MLP falls from $85.0$ to $32.0$ on PushT without it.

\myparagraph{Tying the readout to the port.}
To test the tie between readout and port, we keep the full recipe, including the action target and the learned port $G_\theta$ in the forward step, and replace only $G_\theta^{\top}$ in the readout \eqref{eq:pic-decode} by $Q^{\top}$ for a fixed random orthonormal $Q\in\R^{r\times d_a}$. The identity \eqref{eq:pic-residual} then fails. The residual gains a term that vanishes only when the port coincides with $Q$ (Appendix~\ref{app:pic}), which competes with the state-dependent orientation that prediction asks of the port. This untied readout reaches $83.7$ percent, against $75.0$ without PIC and $93.3$ with the tied readout. An action readout through a fixed orthonormal map therefore already helps, and tying it to the port, which turns it into a pure reweighting of rollout error, accounts for the other half of the gain.

\myparagraph{Prior type and placement.}
Moving the Bures-Wasserstein prior from the code $h$ onto the scored state $s$ lowers PushT success from $93.3$ to $86.3$, which supports regularizing perception and letting the state inherit. Replacing Bures-Wasserstein by SIGReg on $h$ reaches $36.7$ percent at the same epoch $10$, consistent with \eqref{eq:forces}. This is a statement about SIGReg inside H-JEPA, where the prior alone must hold open the code directions outside the slice. It is not a statement about LeWM, where SIGReg acts on the same coordinates as the prediction loss.

\myparagraph{Training at the planning horizon.}
Table~\ref{tab:horizon} (PushT, $r{=}192$, $N{=}50$, epoch $10$) shows that one-step prediction does not compose into a plannable rollout, with or without one-step PIC ($12$ and $34$ percent). Matching both horizons at three or five steps reaches $94$ percent. Cutting PIC to one step while keeping $K_p{=}5$ gives $90$ percent. We therefore set $K_p=K_{\pic}$ to the planning horizon.

\myparagraph{Rank is the capacity knob.}
Table~\ref{tab:rank} shows that the useful control dimension is task dependent (bold: rank used in Table~\ref{tab:main}, parentheses: larger evaluation). Two-Room saturates at every tested rank, Reacher and OGB-Cube degrade at full rank, and PushT improves up to $r{=}D$. At $r=D$ the slice compresses nothing but still separates the scored coordinates from the regularized ones. The selected ranks are capacity choices for the planning metric and not estimates of physical state dimension. An audit of the CEM cost on the $r{=}64$ PushT checkpoint makes this concrete. On the candidate sets scored during evaluation, we fit a linear map from the probed physical variables of Appendix~\ref{app:probe} to the terminal state, split every terminal state into that reconstruction and the orthogonal remainder, and rank candidates by the terminal cost computed from each part alone. The remainder alone has Spearman correlation $0.997$ with the full candidate ranking and recovers $89$ percent of the CEM elites, against $0.66$ and $38$ percent for the physical part. Planning therefore uses predictive directions beyond the variables exposed by standard physical probes.

Appendix~\ref{app:history} reports the history ablation that led to the parameter-free phase, and Appendix~\ref{app:training} the training curves.

\section{Limitations}
\label{sec:limits}

H-JEPA uses mechanics as a structural bias and makes no claim of latent system identification. The learned coordinates are free to rotate around the planning objective, and the audit in Appendix~\ref{app:ph_audit} confirms that component-level physical alignment is task dependent. The explicit Euler step does not preserve the energy properties of the continuous-time field, which our short-horizon planner did not require. The orthonormal port fixes the singular values of the instantaneous action map. This removes a scale ambiguity and makes PIC well conditioned, but it gives up state-dependent action gain. The state rank is selected per task and not learned, and at $r=D$ the state has one null direction (\S\ref{sec:bottleneck}). The inherited covariance is only as isotropic as the trained code, and the measured spectra in Appendix~\ref{app:geometry} remain spread around one. Finally, our evidence is limited to offline pixel control in simulation with short-horizon CEM planning. Longer horizons, real observations, and online data collection remain to be tested.

\section{Conclusion}
\label{sec:conclusion}
We introduced H-JEPA, a world model that learns to plan from pixels without reconstruction by separating perceptual representation learning from the state used for control. A Bures-Wasserstein prior regularizes the perceptual code toward isotropy, while a fixed orthonormal projection transfers its covariance geometry to a control state with task-dependent dimension. A phase-conditioned port-Hamiltonian predictor structures the latent dynamics, and tied port-inverse consistency yields an exact reweighting of rollout error along action-input directions without an additional decoder. Ablations support the benefits of structured dynamics and, on PushT, regularizing perception and tying the inverse readout to the action port. H-JEPA matches or exceeds all evaluated baselines across four pixel-based control benchmarks. Together, these results establish covariance inheritance and structured, action-sensitive prediction as a promising combination for learning representations that support effective planning.


\subsection*{Ethics statement}

This work studies representation learning and planning in standard simulated control benchmarks. It involves no human subjects, no personal data, and no dataset releases beyond configurations for public benchmarks. Although the present experiments are confined to simulation, deploying learned world models on physical systems would require application-specific validation of safety, robustness, and failure behavior.

\subsection*{Reproducibility statement}

Section~\ref{sec:method} specifies the model, training objective, and planning procedure, with \eqref{eq:iso}, \eqref{eq:step}, \eqref{eq:pred}, and \eqref{eq:pic} defining the distributional objective, dynamics update, multi-step prediction loss, and PIC loss. Appendix~\ref{app:pic} derives the PIC identity. Default hyperparameters and the per-task checkpoint epochs are listed in Appendix~\ref{app:config}. The probing protocol follows \citet{deltajepa2026} and \citet{lewm2026} and is restated in Appendix~\ref{app:probe}. Predictor diagnostics are in Appendix~\ref{app:ph_audit}, the geometry and collapse analyses in Appendices~\ref{app:geometry} and~\ref{app:collapse}, and optimization traces in Appendix~\ref{app:training}. An anonymized code release with training configurations for all reported runs is included as supplementary material.

\section*{Acknowlegments}
This work was supported by a Tel Aviv University Center for AI and Data Science (TAD) grant
and by Len Blavatnik and the Blavatnik Family
foundation. This research was also supported by
the Ministry of Innovation, Science \& Technology,
Israel (1001576154) and the Michael J. Fox Foundation (MJFF-022407). The contribution of Tamim Zoabi is part of a  PhD thesis research conducted at Tel Aviv University.

\bibliographystyle{iclr2027_conference}
\bibliography{references}

@inproceedings{agrawal2016learning,
 author = {Agrawal, Pulkit and Nair, Ashvin V and Abbeel, Pieter and Malik, Jitendra and Levine, Sergey},
 booktitle = {Advances in Neural Information Processing Systems},
 editor = {D. Lee and M. Sugiyama and U. Luxburg and I. Guyon and R. Garnett},
 pages = {},
 publisher = {Curran Associates, Inc.},
 title = {Learning to Poke by Poking: Experiential Learning of Intuitive Physics},
 url = {https://proceedings.neurips.cc/paper_files/paper/2016/file/c203d8a151612acf12457e4d67635a95-Paper.pdf},
 volume = {29},
 year = {2016}
}

@inproceedings{alemi2017deep,
title={Deep Variational Information Bottleneck},
author={Alexander A. Alemi and Ian Fischer and Joshua V. Dillon and Kevin Murphy},
booktitle={International Conference on Learning Representations},
year={2017},
url={https://openreview.net/forum?id=HyxQzBceg}
}

@inproceedings{assran2023ijepa,
author = { Assran, Mahmoud and Duval, Quentin and Misra, Ishan and Bojanowski, Piotr and Vincent, Pascal and Rabbat, Michael and LeCun, Yann and Ballas, Nicolas },
booktitle = { 2023 IEEE/CVF Conference on Computer Vision and Pattern Recognition (CVPR) },
title = {{ Self-Supervised Learning from Images with a Joint-Embedding Predictive Architecture }},
year = {2023},
volume = {},
ISSN = {},
pages = {15619-15629},
doi = {10.1109/CVPR52729.2023.01499},
url = {https://doi.ieeecomputersociety.org/10.1109/CVPR52729.2023.01499},
publisher = {IEEE Computer Society},
address = {Los Alamitos, CA, USA},
month =Jun}

@misc{balestriero2025lejepa,
      title={LeJEPA: Provable and Scalable Self-Supervised Learning Without the Heuristics}, 
      author={Randall Balestriero and Yann LeCun},
      year={2025},
      eprint={2511.08544},
      archivePrefix={arXiv},
      primaryClass={cs.LG},
      url={https://arxiv.org/abs/2511.08544}, 
}

@inproceedings{bardes2022vicreg,
title={{VICR}eg: Variance-Invariance-Covariance Regularization for Self-Supervised Learning},
author={Adrien Bardes and Jean Ponce and Yann LeCun},
booktitle={International Conference on Learning Representations},
year={2022},
url={https://openreview.net/forum?id=xm6YD62D1Ub}
}

@article{bardes2024vjepa,
title={Revisiting Feature Prediction for Learning Visual Representations from Video},
author={Adrien Bardes and Quentin Garrido and Jean Ponce and Xinlei Chen and Michael Rabbat and Yann LeCun and Mido Assran and Nicolas Ballas},
journal={Transactions on Machine Learning Research},
issn={2835-8856},
year={2024},
url={https://openreview.net/forum?id=QaCCuDfBk2},
note={Featured Certification}
}

@article{bhatia2019bures,
title = {On the Bures–Wasserstein distance between positive definite matrices},
journal = {Expositiones Mathematicae},
volume = {37},
number = {2},
pages = {165-191},
year = {2019},
issn = {0723-0869},
doi = {https://doi.org/10.1016/j.exmath.2018.01.002},
url = {https://www.sciencedirect.com/science/article/pii/S0723086918300021},
author = {Rajendra Bhatia and Tanvi Jain and Yongdo Lim}
}

@inproceedings{chen2021nsf,
 author = {Chen, Yuhan and Matsubara, Takashi and Yaguchi, Takaharu},
 booktitle = {Advances in Neural Information Processing Systems},
 editor = {M. Ranzato and A. Beygelzimer and Y. Dauphin and P.S. Liang and J. Wortman Vaughan},
 pages = {16659--16670},
 publisher = {Curran Associates, Inc.},
 title = {Neural Symplectic Form: Learning Hamiltonian Equations on General Coordinate Systems},
 url = {https://proceedings.neurips.cc/paper_files/paper/2021/file/8b519f198dd26772e3e82874826b04aa-Paper.pdf},
 volume = {34},
 year = {2021}
}

@article{chi2023diffusion,
author = {Cheng Chi and Zhenjia Xu and Siyuan Feng and Eric Cousineau and Yilun Du and Benjamin Burchfiel and Russ Tedrake and Shuran Song},
title ={Diffusion policy: Visuomotor policy learning via action diffusion},

journal = {The International Journal of Robotics Research},
volume = {44},
number = {10-11},
pages = {1684-1704},
year = {2025},
doi = {10.1177/02783649241273668},

URL = { 
    
        https://doi.org/10.1177/02783649241273668
    
    

},
eprint = { 
    
        https://doi.org/10.1177/02783649241273668
    
    

}
}

@inproceedings{dosovitskiy2021vit,
title={An Image is Worth 16x16 Words: Transformers for Image Recognition at Scale},
author={Alexey Dosovitskiy and Lucas Beyer and Alexander Kolesnikov and Dirk Weissenborn and Xiaohua Zhai and Thomas Unterthiner and Mostafa Dehghani and Matthias Minderer and Georg Heigold and Sylvain Gelly and Jakob Uszkoreit and Neil Houlsby},
booktitle={International Conference on Learning Representations},
year={2021},
url={https://openreview.net/forum?id=YicbFdNTTy}
}

@inproceedings{finzi2020weak,
 author = {Course, Kevin and Evans, Trefor and Nair, Prasanth},
 booktitle = {Advances in Neural Information Processing Systems},
 editor = {H. Larochelle and M. Ranzato and R. Hadsell and M.F. Balcan and H. Lin},
 pages = {18716--18726},
 publisher = {Curran Associates, Inc.},
 title = {Weak Form Generalized Hamiltonian Learning},
 url = {https://proceedings.neurips.cc/paper_files/paper/2020/file/d93c96e6a23fff65b91b900aaa541998-Paper.pdf},
 volume = {33},
 year = {2020}
}

@inproceedings{gelada2019deepmdp,
  title = 	 {{D}eep{MDP}: Learning Continuous Latent Space Models for Representation Learning},
  author =       {Gelada, Carles and Kumar, Saurabh and Buckman, Jacob and Nachum, Ofir and Bellemare, Marc G.},
  booktitle = 	 {Proceedings of the 36th International Conference on Machine Learning},
  pages = 	 {2170--2179},
  year = 	 {2019},
  editor = 	 {Chaudhuri, Kamalika and Salakhutdinov, Ruslan},
  volume = 	 {97},
  series = 	 {Proceedings of Machine Learning Research},
  month = 	 {09--15 Jun},
  publisher =    {PMLR},
  url = 	 {https://proceedings.mlr.press/v97/gelada19a.html}
}

@article{gelbrich1990,
author = {Gelbrich, Matthias},
title = {On a Formula for the L2 Wasserstein Metric between Measures on Euclidean and Hilbert Spaces},
journal = {Mathematische Nachrichten},
volume = {147},
number = {1},
pages = {185-203},
doi = {https://doi.org/10.1002/mana.19901470121},
url = {https://onlinelibrary.wiley.com/doi/abs/10.1002/mana.19901470121},
eprint = {https://onlinelibrary.wiley.com/doi/pdf/10.1002/mana.19901470121},
year = {1990}
}

@inproceedings{greydanus2019hnn,
 author = {Greydanus, Samuel and Dzamba, Misko and Yosinski, Jason},
 booktitle = {Advances in Neural Information Processing Systems},
 editor = {H. Wallach and H. Larochelle and A. Beygelzimer and F. d\textquotesingle Alch\'{e}-Buc and E. Fox and R. Garnett},
 pages = {},
 publisher = {Curran Associates, Inc.},
 title = {Hamiltonian Neural Networks},
 url = {https://proceedings.neurips.cc/paper_files/paper/2019/file/26cd8ecadce0d4efd6cc8a8725cbd1f8-Paper.pdf},
 volume = {32},
 year = {2019}
}

@inproceedings{grill2020byol,
 author = {Grill, Jean-Bastien and Strub, Florian and Altch\'{e}, Florent and Tallec, Corentin and Richemond, Pierre and Buchatskaya, Elena and Doersch, Carl and Avila Pires, Bernardo and Guo, Zhaohan and Gheshlaghi Azar, Mohammad and Piot, Bilal and kavukcuoglu, koray and Munos, Remi and Valko, Michal},
 booktitle = {Advances in Neural Information Processing Systems},
 editor = {H. Larochelle and M. Ranzato and R. Hadsell and M.F. Balcan and H. Lin},
 pages = {21271--21284},
 publisher = {Curran Associates, Inc.},
 title = {Bootstrap Your Own Latent - A New Approach to Self-Supervised Learning},
 url = {https://proceedings.neurips.cc/paper_files/paper/2020/file/f3ada80d5c4ee70142b17b8192b2958e-Paper.pdf},
 volume = {33},
 year = {2020}
}

@misc{pit2026,
      title={Rethinking Weight Tying: Pseudo-Inverse Tying for LM Stable Training and Updates}, 
      author={Jian Gu and Aldeida Aleti and Chunyang Chen and Hongyu Zhang},
      year={2026},
      eprint={2602.04556},
      archivePrefix={arXiv},
      primaryClass={cs.CL},
      url={https://arxiv.org/abs/2602.04556}, 
}

@inproceedings{ha2018world,
 author = {Ha, David and Schmidhuber, J\"{u}rgen},
 booktitle = {Advances in Neural Information Processing Systems},
 editor = {S. Bengio and H. Wallach and H. Larochelle and K. Grauman and N. Cesa-Bianchi and R. Garnett},
 pages = {},
 publisher = {Curran Associates, Inc.},
 title = {Recurrent World Models Facilitate Policy Evolution},
 url = {https://proceedings.neurips.cc/paper_files/paper/2018/file/2de5d16682c3c35007e4e92982f1a2ba-Paper.pdf},
 volume = {31},
 year = {2018}
}

@inproceedings{hafner2019planet,
  title = 	 {Learning Latent Dynamics for Planning from Pixels},
  author =       {Hafner, Danijar and Lillicrap, Timothy and Fischer, Ian and Villegas, Ruben and Ha, David and Lee, Honglak and Davidson, James},
  booktitle = 	 {Proceedings of the 36th International Conference on Machine Learning},
  pages = 	 {2555--2565},
  year = 	 {2019},
  editor = 	 {Chaudhuri, Kamalika and Salakhutdinov, Ruslan},
  volume = 	 {97},
  series = 	 {Proceedings of Machine Learning Research},
  month = 	 {09--15 Jun},
  publisher =    {PMLR},
  url = 	 {https://proceedings.mlr.press/v97/hafner19a.html}
}

@misc{hafner2020dreamer,
      title={Dream to Control: Learning Behaviors by Latent Imagination}, 
      author={Danijar Hafner and Timothy Lillicrap and Jimmy Ba and Mohammad Norouzi},
      year={2020},
      eprint={1912.01603},
      archivePrefix={arXiv},
      primaryClass={cs.LG},
      url={https://arxiv.org/abs/1912.01603}, 
}

@misc{smwm2026,
      title={Sensorimotor World Models: Perception for Action via Inverse Dynamics}, 
      author={Petr Ivashkov and Randall Balestriero and Bernhard Schölkopf},
      year={2026},
      eprint={2606.20104},
      archivePrefix={arXiv},
      primaryClass={cs.LG},
      url={https://arxiv.org/abs/2606.20104}, 
}

@inproceedings{karl2017dvbf,
title={Deep Variational Bayes Filters: Unsupervised Learning of State Space Models from Raw Data},
author={Maximilian Karl and Maximilian Soelch and Justin Bayer and Patrick van der Smagt},
booktitle={International Conference on Learning Representations},
year={2017},
url={https://openreview.net/forum?id=HyTqHL5xg}
}

@article{lamb2022acstate,
title={Guaranteed Discovery of Control-Endogenous Latent States with Multi-Step Inverse Models},
author={Alex Lamb and Riashat Islam and Yonathan Efroni and Aniket Rajiv Didolkar and Dipendra Misra and Dylan J Foster and Lekan P Molu and Rajan Chari and Akshay Krishnamurthy and John Langford},
journal={Transactions on Machine Learning Research},
issn={2835-8856},
year={2023},
url={https://openreview.net/forum?id=TNocbXm5MZ},
note={}
}

@article{lecun2022path,
  author  = {LeCun, Yann},
  title   = {A path towards autonomous machine intelligence},
  journal = {OpenReview preprint},
  year    = {2022},
}

@inproceedings{
levine2024acdf,
title={Multistep Inverse Is Not All You Need},
author={Alexander Levine and Peter Stone and Amy Zhang},
booktitle={Reinforcement Learning Conference},
year={2024},
url={https://openreview.net/forum?id=xyrgG4rsqY}
}

@inproceedings{littman2001psr,
 author = {Littman, Michael and Sutton, Richard S},
 booktitle = {Advances in Neural Information Processing Systems},
 editor = {T. Dietterich and S. Becker and Z. Ghahramani},
 pages = {},
 publisher = {MIT Press},
 title = {Predictive Representations of State},
 url = {https://proceedings.neurips.cc/paper_files/paper/2001/file/1e4d36177d71bbb3558e43af9577d70e-Paper.pdf},
 volume = {14},
 year = {2001}
}

@inproceedings{loshchilov2019adamw,
title={Decoupled Weight Decay Regularization},
author={Ilya Loshchilov and Frank Hutter},
booktitle={International Conference on Learning Representations},
year={2019},
url={https://openreview.net/forum?id=Bkg6RiCqY7},
}

@misc{lewm2026,
      title={LeWorldModel: Stable End-to-End Joint-Embedding Predictive Architecture from Pixels}, 
      author={Lucas Maes and Quentin Le Lidec and Damien Scieur and Yann LeCun and Randall Balestriero},
      year={2026},
      eprint={2603.19312},
      archivePrefix={arXiv},
      primaryClass={cs.LG},
      url={https://arxiv.org/abs/2603.19312}, 
}

@article{marchenko1967,
doi = {10.1070/SM1967v001n04ABEH001994},
url = {https://doi.org/10.1070/SM1967v001n04ABEH001994},
year = {1967},
month = {apr},
publisher = {},
volume = {1},
number = {4},
pages = {457},
author = {V A Marčenko and L A Pastur},
title = {DISTRIBUTION OF EIGENVALUES FOR SOME SETS OF RANDOM MATRICES},
journal = {Mathematics of the USSR-Sbornik}
}

@misc{mishkin2016lsuv,
      title={All you need is a good init}, 
      author={Dmytro Mishkin and Jiri Matas},
      year={2016},
      eprint={1511.06422},
      archivePrefix={arXiv},
      primaryClass={cs.LG},
      url={https://arxiv.org/abs/1511.06422}, 
}

@article{moore1981balanced,
  author={Moore, B.},
  journal={IEEE Transactions on Automatic Control}, 
  title={Principal component analysis in linear systems: Controllability, observability, and model reduction}, 
  year={1981},
  volume={26},
  number={1},
  pages={17-32},
  doi={10.1109/TAC.1981.1102568}}

@book{murray1994mathematical,
  author    = {Murray, Richard M. and Li, Zexiang and Sastry, S. Shankar},
  title     = {A Mathematical Introduction to Robotic Manipulation},
  publisher = {CRC Press},
  year      = {1994},
}

@inproceedings{park2025ogbench,
 author = {Park, Seohong and Frans, Kevin and Eysenbach, Benjamin and Levine, Sergey},
 booktitle = {International Conference on Learning Representations},
 editor = {Y. Yue and A. Garg and N. Peng and F. Sha and R. Yu},
 pages = {94937--94982},
 title = {OGBench: Benchmarking Offline Goal-Conditioned RL},
 url = {https://proceedings.iclr.cc/paper_files/paper/2025/file/ecd92623ac899357312aaa8915853699-Paper-Conference.pdf},
 volume = {2025},
 year = {2025}
}

@inproceedings{pathak2017curiosity,
  title = 	 {Curiosity-driven Exploration by Self-supervised Prediction},
  author =       {Deepak Pathak and Pulkit Agrawal and Alexei A. Efros and Trevor Darrell},
  booktitle = 	 {Proceedings of the 34th International Conference on Machine Learning},
  pages = 	 {2778--2787},
  year = 	 {2017},
  editor = 	 {Precup, Doina and Teh, Yee Whye},
  volume = 	 {70},
  series = 	 {Proceedings of Machine Learning Research},
  month = 	 {06--11 Aug},
  publisher =    {PMLR},
  url = 	 {https://proceedings.mlr.press/v70/pathak17a.html}
}

@inproceedings{press2017tying,
    title = "Using the Output Embedding to Improve Language Models",
    author = "Press, Ofir  and
      Wolf, Lior",
    editor = "Lapata, Mirella  and
      Blunsom, Phil  and
      Koller, Alexander",
    booktitle = "Proceedings of the 15th Conference of the {E}uropean Chapter of the Association for Computational Linguistics: Volume 2, Short Papers",
    month = apr,
    year = "2017",
    address = "Valencia, Spain",
    publisher = "Association for Computational Linguistics",
    url = "https://aclanthology.org/E17-2025/",
    pages = "157--163"
}

@article{rubinstein1999cem,
journal={Methodology and Computing in Applied Probability},
author={Reuven Rubinstein},
title={The Cross-Entropy Method for Combinatorial and Continuous Optimization},
year={1999},
month={September},
pages={127-190},
volume={1},
number={2},
doi={10.1023/A:1010091220143},
url={https://ideas.repec.org/a/spr/metcap/v1y1999i2d10.1023_a1010091220143.html},
}

@inproceedings{pldm2025,
 author = {Sobal, Uladzislau and Zhang, Wancong and Cho, Kyunghyun and Balestriero, Randall and Rudner, Tim G. J. and LeCun, Yann},
 booktitle = {Advances in Neural Information Processing Systems},
 doi = {10.52202/085713-1465},
 editor = {D. Belgrave and C. Zhang and H. Lin and R. Pascanu and P. Koniusz and M. Ghassemi and N. Chen},
 pages = {43905--43941},
 publisher = {Curran Associates, Inc.},
 title = {Learning from Reward-Free Offline Data: A Case for Planning with Latent Dynamics Models},
 url = {https://proceedings.neurips.cc/paper_files/paper/2025/file/3e7cf447f21cd11c846463affefce665-Paper-Conference.pdf},
 volume = {38, Main Conference},
 year = {2025}
}

@misc{tassa2018dmc,
      title={DeepMind Control Suite}, 
      author={Yuval Tassa and Yotam Doron and Alistair Muldal and Tom Erez and Yazhe Li and Diego de Las Casas and David Budden and Abbas Abdolmaleki and Josh Merel and Andrew Lefrancq and Timothy Lillicrap and Martin Riedmiller},
      year={2018},
      eprint={1801.00690},
      archivePrefix={arXiv},
      primaryClass={cs.AI},
      url={https://arxiv.org/abs/1801.00690}, 
}

@misc{tishby1999information,
      title={The information bottleneck method}, 
      author={Naftali Tishby and Fernando C. Pereira and William Bialek},
      year={2000},
      eprint={physics/0004057},
      archivePrefix={arXiv},
      primaryClass={physics.data-an},
      url={https://arxiv.org/abs/physics/0004057}, 
}

@inproceedings{toth2020hgn,
title={Hamiltonian Generative Networks},
author={Peter Toth and Danilo J. Rezende and Andrew Jaegle and Sébastien Racanière and Aleksandar Botev and Irina Higgins},
booktitle={International Conference on Learning Representations},
year={2020},
url={https://openreview.net/forum?id=HJenn6VFvB}
}

@article{vanderschaft2014ph,
    author = {van der Schaft, Arjan and Jeltsema, Dimitri},
    title = {Port-Hamiltonian Systems Theory: An Introductory Overview},
    journal = {Foundations and Trends in Systems and Control},
    volume = {1},
    number = {2-3},
    pages = {173-378},
    year = {2014},
    month = {06},
    issn = {2325-6818},
    doi = {10.1561/2600000002},
    url = {https://doi.org/10.1561/2600000002},
    eprint = {https://www.emerald.com/ftsys/article-pdf/1/2-3/173/11146089/2600000002en.pdf},
}

@inproceedings{watter2015embed,
 author = {Watter, Manuel and Springenberg, Jost and Boedecker, Joschka and Riedmiller, Martin},
 booktitle = {Advances in Neural Information Processing Systems},
 editor = {C. Cortes and N. Lawrence and D. Lee and M. Sugiyama and R. Garnett},
 pages = {},
 publisher = {Curran Associates, Inc.},
 title = {Embed to Control: A Locally Linear Latent Dynamics Model for Control from Raw Images},
 url = {https://proceedings.neurips.cc/paper_files/paper/2015/file/a1afc58c6ca9540d057299ec3016d726-Paper.pdf},
 volume = {28},
 year = {2015}
}

@inproceedings{zbontar2021barlow,
  title = 	 {Barlow Twins: Self-Supervised Learning via Redundancy Reduction},
  author =       {Zbontar, Jure and Jing, Li and Misra, Ishan and LeCun, Yann and Deny, Stephane},
  booktitle = 	 {Proceedings of the 38th International Conference on Machine Learning},
  pages = 	 {12310--12320},
  year = 	 {2021},
  editor = 	 {Meila, Marina and Zhang, Tong},
  volume = 	 {139},
  series = 	 {Proceedings of Machine Learning Research},
  month = 	 {18--24 Jul},
  publisher =    {PMLR},
  url = 	 {https://proceedings.mlr.press/v139/zbontar21a.html}
}

@inproceedings{zhang2021bisim,
title={Learning Invariant Representations for Reinforcement Learning without Reconstruction},
author={Amy Zhang and Rowan Thomas McAllister and Roberto Calandra and Yarin Gal and Sergey Levine},
booktitle={International Conference on Learning Representations},
year={2021},
url={https://openreview.net/forum?id=-2FCwDKRREu}
}

@misc{deltajepa2026,
      title={Delta-JEPA: Learning Action-Sensitive World Models via Latent Difference Decoding}, 
      author={Zhenghao Zhang and Yuanxiang Wang and Zhenyu Guan and Yujia Yang and Bingkang Shi and Tianyu Zong and Hongzhu Yi and Guoqing Chao and Xingchen Chen and Tiankun Yang and Chenxi Bao and Tao Yu and Jingjing Zhou and Jungang Xu},
      year={2026},
      eprint={2606.31232},
      archivePrefix={arXiv},
      primaryClass={cs.AI},
      url={https://arxiv.org/abs/2606.31232}, 
}

@misc{subjepa2026,
      title={Sub-JEPA: Subspace Gaussian Regularization for Stable End-to-End World Models}, 
      author={Kai Zhao and Dongliang Nie and Yuchen Lin and Zhehan Luo and Yixiao Gu and Deng-Ping Fan and Dan Zeng},
      year={2026},
      eprint={2605.09241},
      archivePrefix={arXiv},
      primaryClass={cs.LG},
      url={https://arxiv.org/abs/2605.09241}, 
}

@inproceedings{zhong2020symplectic,
title={Symplectic ODE-Net: Learning Hamiltonian Dynamics with Control},
author={Yaofeng Desmond Zhong and Biswadip Dey and Amit Chakraborty},
booktitle={International Conference on Learning Representations},
year={2020},
url={https://openreview.net/forum?id=ryxmb1rKDS}
}

@inproceedings{zhou2024dinowm,

title={{DINO}-{WM}: World Models on Pre-trained Visual Features enable Zero-shot Planning},
author={Gaoyue Zhou and Hengkai Pan and Yann LeCun and Lerrel Pinto},
booktitle={Forty-second International Conference on Machine Learning},
year={2025},
url={https://openreview.net/forum?id=D5RNACOZEI}
}

@article{epps1983test,
    author = {EPPS, T. W. and PULLEY, LAWRENCE B.},
    title = {A test for normality based on the empirical characteristic function},
    journal = {Biometrika},
    volume = {70},
    number = {3},
    pages = {723-726},
    year = {1983},
    month = {12},
    issn = {0006-3444},
    doi = {10.1093/biomet/70.3.723},
    url = {https://doi.org/10.1093/biomet/70.3.723},
    eprint = {https://academic.oup.com/biomet/article-pdf/70/3/723/687464/70-3-723.pdf},
}

\appendix

\section{Architecture overview}
\label{app:arch}

\begin{figure}[!h]
\centering
\includegraphics[width=0.995\linewidth]{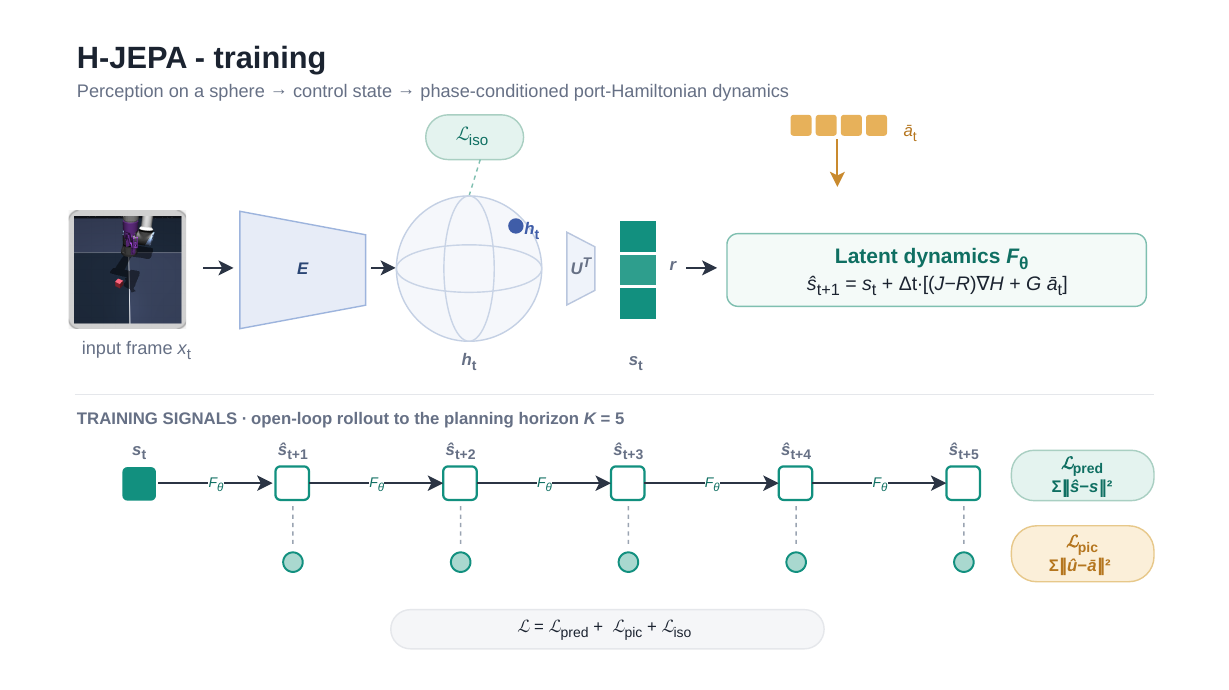}
\caption{\textbf{H-JEPA.} The Bures-Wasserstein prior $\mathcal{L}_{\iso}$ acts only on the spherical code $h_t$. A fixed orthonormal slice $U^{\top}$ yields the control state $s_t$. The phase-conditioned port-Hamiltonian predictor $\mathcal{F}_\theta$ is supervised open-loop ($\mathcal{L}_{\pred}$), PIC reads actions back through the port transpose ($\mathcal{L}_{\pic}$), and planning scores $s$ alone.}
\label{fig:arch}
\end{figure}

Figure~\ref{fig:arch} summarizes the training-time model: perception on a sphere, the fixed orthonormal slice, the port-Hamiltonian predictor, and the three losses.

\section{Derivation of the PIC identity}
\label{app:pic}

Fix a rollout step $i$ and abbreviate $c=\hat c_{t+i}$, $s=\hat s_{t+i}$, $v=\hat v_{t+i}$, and $\bar a=\bar a_{t+i}$. The model's one-step prediction from the rollout state is
\begin{equation}
\hat s_{t+i+1}
= s + \Delta t\Bigl[\bigl(J-R_\theta(c)\bigr)\nabla_s H_\theta(s,v) + G_\theta(c)\,\bar a\Bigr].
\end{equation}
Substituting the readout of \eqref{eq:pic-decode} and using $G_\theta(c)^{\top}G_\theta(c)=I$,
\begin{align}
\hat u_{t+i}-\bar a
&= G_\theta(c)^{\top}\Bigl[\tfrac{s_{t+i+1}-s}{\Delta t}-\bigl(J-R_\theta(c)\bigr)\nabla_s H_\theta(s,v)\Bigr] - G_\theta(c)^{\top}G_\theta(c)\,\bar a\\
&= \tfrac{1}{\Delta t}\,G_\theta(c)^{\top}\Bigl[s_{t+i+1}-s-\Delta t\bigl(J-R_\theta(c)\bigr)\nabla_s H_\theta(s,v)-\Delta t\,G_\theta(c)\,\bar a\Bigr]\\
&= \tfrac{1}{\Delta t}\,G_\theta(c)^{\top}\bigl(s_{t+i+1}-\hat s_{t+i+1}\bigr)
= \tfrac{1}{\Delta t}\,G_\theta(c)^{\top}\delta_{t+i+1},
\end{align}
which is \eqref{eq:pic-residual}. The PIC residual is the projection of the open-loop state error onto the port subspace, scaled by $1/\Delta t$. Summing $\tfrac1r\|\delta\|_2^2$ and $\tfrac{1}{d_a\Delta t^2}\|G^{\top}\delta\|_2^2$ over the same steps gives the quadratic form in \eqref{eq:pic-metric}, which weights error along the $d_a$ port directions by a factor $1+r/(d_a\Delta t^{2})$ relative to the remaining $r-d_a$ directions. With $\Delta t=1$ this factor is $17$ on Two-Room, $33$ on Reacher, $97$ on PushT at $r=192$, and $7.4$ on OGB-Cube.

The identity does not remove the action readout. On a transition that the model predicts exactly, the transpose map returns $\hat u=\bar a$. Off such transitions the readout is an orthogonal projection of the residual displacement, so it never amplifies noise, its magnitude is bounded by the transition energy, and $\mathcal{L}_{\pic}$ is of order one at initialization in normalized units. The identity also explains the control in \S\ref{sec:abl}, where the readout uses a fixed random orthonormal $Q^{\top}$ in place of $G_\theta(c)^{\top}$ while the forward step keeps $G_\theta$. The same substitution now gives
\begin{equation}
\hat u_{t+i}-\bar a
= \tfrac{1}{\Delta t}\,Q^{\top}\delta_{t+i+1}+\bigl(Q^{\top}G_\theta(c)-I\bigr)\bar a .
\end{equation}
The first term is again a projected rollout error. The second is new and vanishes only if $G_\theta(c)=Q$ for every visited phase, so the untied loss pulls the port toward a fixed frame while $\mathcal{L}_{\pred}$ asks it to follow the state. With the tied readout this term is identically zero and the port is left free. The control recovers about half of the gain ($75.0$ to $83.7$, against $93.3$ with the tied readout), and the tie supplies the rest. Weighting error along the directions in which actions enter is qualitatively related to the controllability side of balanced realizations \citep{moore1981balanced}, with predictive accuracy playing the analogous role for observability.

\section{Default configuration and training details}
\label{app:config}

\begin{table}[h]
\centering
\caption{Default H-JEPA configuration. Task-specific entries are shown explicitly for rank, action dimension, and checkpoint epoch.}
\label{tab:config}
\small
\begin{tabular}{@{}lp{9.6cm}@{}}
\toprule
Quantity & Value \\
\midrule
Perceptual dimension $D$ & $192$ \\
Encoder & ViT-T/14 from scratch, MLP projector $192\to2048\to192$ \\
Output normalization & affine-free LayerNorm, $\|h\|=\sqrt{D}$, per sample \\
Projection $U$ & fixed random orthonormal, seeded QR \\
History and phase & $2$ frames, $c=[s,v]$ with $v=s-s_{\mathrm{prev}}$, left-pad $v_0=0$ \\
Net action dimension $d_a$ & $2$ on Two-Room, Reacher, PushT, $5$ on OGB-Cube \\
Port constraint & Stiefel, $G^{\top}G=I$ via differentiable QR \\
Dissipation & $R=LL^{\top}$ with lower-triangular $L$, seeded so $R\approx(\rho_{\mathrm{init}}/r)I$, $\rho_{\mathrm{init}}=10^{-2}$ \\
Dynamics MLPs & depth $3$, width $256$, GELU, no dropout \\
$H_\theta$ head initialization & $\mathcal{N}(0,10^{-3})$ so passive drift starts near zero \\
Step size $\Delta t$ & $1.0$ \\
Horizons & $K_p=5$ (equal to the CEM horizon), $K_{\pic}=5$ \\
Window length $T$ & $7$ \\
Optimizer & AdamW, lr $5\times10^{-5}$, weight decay $10^{-3}$, bf16, grad clipping \\
Training length and seeds & $10$ epochs on every task, checkpoint epoch $8/6/10/3$ for Two-Room/Reacher/PushT/OGB-Cube selected on a held-out pilot seed, reported seeds $\{3072,7,11\}$ \\
Baseline protocol & Delta-JEPA evaluation protocol \citep{deltajepa2026}, with $50$ training epochs for Delta-JEPA \\
CEM & horizon $5$, receding $5$, $300$ samples, $30$ iters PushT, $10$ elsewhere \\
Evaluation & $N=500$ main, lower-budget ablations use $N=50$ or $N=300$ as stated \\
Diagnostic probe & linear head on detached displacements, logging only \\
\bottomrule
\end{tabular}
\end{table}

Table~\ref{tab:config} lists the defaults. The three losses are summed without weights, which places PIC and prediction on comparable scales in normalized action and state units. For context, \citet{deltajepa2026} sweep the weight of their learned action decoder in the same action units and report collapse below $0.1$ with a broad workable range from $1$ to $50$. The two losses are not equivalent, so we treat this as context and not as hyperparameter transfer. The readout itself is two matrix products plus one thin QR factorization per step, with no linear solve and no regularizer, run under bf16 autocast with the factorization in a small fp32 island, as is standard for spectral operations.

\section{Full physical probing tables}
\label{app:probe}

The protocol follows \citet{deltajepa2026} and \citet{lewm2026}: $20{,}000$ frames, trajectory-level train and test split, targets $z$-scored on the train split, an OLS linear probe and a $2{\times}256$ ReLU MLP probe trained for $100$ epochs, and three probe seeds governing MLP initialization and minibatch order (the OLS fit is deterministic given the split), reported as mean $\pm$ std. H-JEPA probes the control state $s=U^{\top}h$ at the task-specific converged checkpoints defined in \S\ref{sec:setup} (PushT seed $11$ at $r{=}192$, Reacher seed $3072$ at $r{=}64$, OGB-Cube seed $3072$ at $r{=}32$), and LeWM and Sub-JEPA use their official projected $192$-d embeddings. Angular and quaternion targets are evaluated in raw coordinates, so wrap-around and sign ambiguity depress correlations for all methods alike, and those rows should be read comparatively. Tables~\ref{tab:probe}, \ref{tab:probe-reacher}, and~\ref{tab:probe-cube} report PushT, Reacher, and OGB-Cube in turn.

\begin{table}[t]
\centering
\caption{Physical latent probing on PushT ($\rho$ is the Pearson correlation).}
\label{tab:probe}
\small
\begin{tabular}{@{}llcccc@{}}
\toprule
Property & Method & Linear MSE $\downarrow$ & Linear $\rho$ $\uparrow$ & MLP MSE $\downarrow$ & MLP $\rho$ $\uparrow$ \\
\midrule
Agent location & H-JEPA & $0.028\pm0.003$ & $0.986$ & $0.008\pm0.003$ & $0.996$ \\
 & LeWM & $0.062\pm0.012$ & $0.968$ & $0.017\pm0.006$ & $0.992$ \\
 & Sub-JEPA & $0.065\pm0.009$ & $0.966$ & $0.016\pm0.004$ & $0.992$ \\
\midrule
Block location & H-JEPA & $0.152\pm0.025$ & $0.926$ & $0.049\pm0.024$ & $0.979$ \\
 & LeWM & $0.038\pm0.011$ & $0.982$ & $0.024\pm0.014$ & $0.990$ \\
 & Sub-JEPA & $0.033\pm0.009$ & $0.984$ & $0.017\pm0.007$ & $0.992$ \\
\midrule
Block angle & H-JEPA & $0.331\pm0.053$ & $0.822$ & $0.199\pm0.024$ & $0.896$ \\
 & LeWM & $0.199\pm0.019$ & $0.898$ & $0.108\pm0.025$ & $0.944$ \\
 & Sub-JEPA & $0.201\pm0.013$ & $0.895$ & $0.072\pm0.015$ & $0.965$ \\
\bottomrule
\end{tabular}
\end{table}

\begin{table}[t]
\centering
\caption{Physical latent probing on Reacher ($\rho$ is the Pearson correlation).}
\label{tab:probe-reacher}
\small
\begin{tabular}{@{}llcccc@{}}
\toprule
Property & Method & Linear MSE $\downarrow$ & Linear $\rho$ $\uparrow$ & MLP MSE $\downarrow$ & MLP $\rho$ $\uparrow$ \\
\midrule
Finger position & H-JEPA & $0.002\pm0.000$ & $0.999$ & $0.000\pm0.000$ & $1.000$ \\
 & LeWM & $0.002\pm0.000$ & $0.999$ & $0.000\pm0.000$ & $1.000$ \\
 & Sub-JEPA & $0.002\pm0.000$ & $0.999$ & $0.000\pm0.000$ & $1.000$ \\
\midrule
Joint position & H-JEPA & $0.601\pm0.125$ & $0.588$ & $0.742\pm0.111$ & $0.562$ \\
 & LeWM & $0.636\pm0.120$ & $0.573$ & $0.738\pm0.116$ & $0.550$ \\
 & Sub-JEPA & $0.639\pm0.125$ & $0.575$ & $0.726\pm0.129$ & $0.559$ \\
\bottomrule
\end{tabular}
\end{table}

\begin{table}[t]
\centering
\caption{Physical latent probing on OGB-Cube ($\rho$ is the Pearson correlation).}
\label{tab:probe-cube}
\small
\begin{tabular}{@{}llcccc@{}}
\toprule
Property & Method & Linear MSE $\downarrow$ & Linear $\rho$ $\uparrow$ & MLP MSE $\downarrow$ & MLP $\rho$ $\uparrow$ \\
\midrule
Joint position & H-JEPA & $0.445\pm0.044$ & $0.670$ & $0.554\pm0.056$ & $0.685$ \\
 & LeWM & $0.579\pm0.060$ & $0.645$ & $0.775\pm0.109$ & $0.647$ \\
 & Sub-JEPA & $0.492\pm0.067$ & $0.676$ & $0.557\pm0.059$ & $0.675$ \\
\midrule
Joint velocity & H-JEPA & $0.740\pm0.027$ & $0.432$ & $0.713\pm0.039$ & $0.541$ \\
 & LeWM & $1.069\pm0.038$ & $0.088$ & $2.674\pm0.236$ & $0.050$ \\
 & Sub-JEPA & $0.988\pm0.032$ & $0.203$ & $1.595\pm0.226$ & $0.101$ \\
\midrule
End-effector position & H-JEPA & $0.007\pm0.000$ & $0.997$ & $0.002\pm0.000$ & $0.999$ \\
 & LeWM & $0.027\pm0.003$ & $0.987$ & $0.020\pm0.001$ & $0.991$ \\
 & Sub-JEPA & $0.025\pm0.002$ & $0.988$ & $0.014\pm0.000$ & $0.993$ \\
\midrule
End-effector yaw & H-JEPA & $1.315\pm0.291$ & $-0.124$ & $2.133\pm0.448$ & $-0.113$ \\
 & LeWM & $2.046\pm0.483$ & $-0.161$ & $2.198\pm0.370$ & $-0.143$ \\
 & Sub-JEPA & $1.575\pm0.411$ & $0.008$ & $2.115\pm0.380$ & $-0.078$ \\
\midrule
Block position & H-JEPA & $0.023\pm0.001$ & $0.989$ & $0.009\pm0.001$ & $0.996$ \\
 & LeWM & $0.011\pm0.001$ & $0.995$ & $0.010\pm0.001$ & $0.995$ \\
 & Sub-JEPA & $0.053\pm0.005$ & $0.975$ & $0.010\pm0.001$ & $0.995$ \\
\midrule
Block quaternion & H-JEPA & $1.352\pm0.439$ & $0.138$ & $1.753\pm0.421$ & $0.103$ \\
 & LeWM & $1.822\pm0.403$ & $-0.010$ & $2.512\pm0.505$ & $-0.035$ \\
 & Sub-JEPA & $1.662\pm0.434$ & $0.021$ & $2.273\pm0.358$ & $-0.010$ \\
\midrule
Block yaw & H-JEPA & $1.142\pm0.233$ & $0.184$ & $1.804\pm0.222$ & $0.024$ \\
 & LeWM & $1.779\pm0.448$ & $-0.025$ & $1.856\pm0.368$ & $-0.050$ \\
 & Sub-JEPA & $1.313\pm0.270$ & $0.208$ & $1.589\pm0.558$ & $0.161$ \\
\bottomrule
\end{tabular}
\end{table}

\section{Interpreting the port-Hamiltonian predictor}
\label{app:ph_audit}

The port-Hamiltonian parameterization is introduced as an inductive bias rather
than an assumption that the learned coordinates recover physical positions,
momenta, or energies. We therefore ask a more limited question: \emph{after
training for planning, does the predictor use its structured components in a
meaningful and task-dependent way?} We audit the best planning checkpoint for
each benchmark on $24{,}576$ held-out expert transitions, evaluating the learned
fields only on encoded states encountered in the data.

\myparagraph{The imposed structure is realized numerically.}
For each transition we decompose the predicted velocity into
\begin{equation}
\dot s
=
\underbrace{\bigl( J - R(c) \bigr) \nabla_s H(s, v)}_{\text{passive flow}}
+
\underbrace{G(c) \bar a}_{\text{control flow}} .
\label{eq:app_ph_decomp}
\end{equation}
The Stiefel constraint is satisfied to numerical precision on all three tasks,
$\| G^\top G - I \|_\infty \approx 4.8 \times 10^{-7}$, while
$R(c) = L(c) L(c)^\top$ remains positive semidefinite by construction. Thus, the
analysis below probes the learned fields within the intended decomposition
and not artifacts of constraint violation.

\myparagraph{The balance of passive and controlled dynamics is task dependent.}
Table~\ref{tab:ph_audit_structure} shows that the predictor does not use the
same decomposition across environments. PushT is strongly
dominated by passive, dissipative flow:
$\| R \nabla H \| / \| J \nabla H \| = 4.35$, while the direct control flow is
only $0.13$ times the passive flow on expert trajectories. Reacher is more
balanced, and OGB-Cube becomes control dominated, with
$\| G \bar a \| / \| \mathrm{passive} \| = 1.53$. The overlap
$\operatorname{tr}(G^\top R G) / \operatorname{tr}(R)$ is nearly zero on PushT
but substantially larger on OGB-Cube, indicating that the learned dissipative and
actuated subspaces themselves adapt to the task.

\begin{table}[t]
\centering
\small
\caption{\textbf{Learned port-Hamiltonian structure on held-out transitions.}
The relative contribution of dissipation, passive dynamics, and direct control
changes markedly across tasks.}
\label{tab:ph_audit_structure}
\begin{tabular}{lcccc}
\toprule
Task &
$\dfrac{\| R \nabla H \|}{\| J \nabla H \|}$ &
$\dfrac{\| G \bar a \|}{\| \mathrm{passive} \|}$ &
$\dfrac{\operatorname{tr}(G^\top R G)}{\operatorname{tr}(R)}$ &
$\operatorname{corr}(\operatorname{tr} R, \| v \|)$ \\[1.2ex]
\midrule
PushT   & 4.35 & 0.13 & 0.001 & 0.48 \\
Reacher & 1.16 & 0.56 & 0.008 & 0.24 \\
OGB-Cube    & 1.11 & 1.53 & 0.184 & 0.13 \\
\bottomrule
\end{tabular}
\end{table}

PushT provides the clearest physical interpretation. Its learned dissipation
increases with latent speed
($\operatorname{corr}(\operatorname{tr} R, \| v \|) = 0.48$), and the
dissipative component dominates the conservative component, consistent with a
representation organized around the strongly dissipative contact dynamics of
pushing. OGB-Cube exhibits the opposite regime: direct actuation is larger than the
passive field and actuation overlaps more strongly with the learned dissipative
subspace. We emphasize that these quantities are measured in learned latent
coordinates and should be interpreted as structural signatures, not estimates of
physical coefficients in SI units.

\myparagraph{When the task permits it, the learned fields align with physical change.}
To test whether the decomposition contains interpretable information beyond its
relative magnitudes, we fit linear ridge probes from each field to the
corresponding physical state increment $\Delta y$. Results are shown in
Table~\ref{tab:ph_audit_alignment}. On PushT, the passive field, control field,
and Hamiltonian gradient each explain approximately $88$ to $90$ percent of the variance
in the physical increment. The control state itself predicts the physical state
with $R^2 = 0.95$. Thus, on this task the latent decomposition is not only
structurally valid but closely aligned with observable physical evolution.

\begin{table}[t]
\centering
\small
\caption{\textbf{Linear physical alignment of the learned predictor.}
$R^2$ is measured on held-out transitions using ridge regression.}
\label{tab:ph_audit_alignment}
\begin{tabular}{lcccc}
\toprule
Task &
Passive $\rightarrow \Delta y$ &
$G \bar a \rightarrow \Delta y$ &
$\nabla H \rightarrow \Delta y$ &
$s \rightarrow y$ \\
\midrule
PushT   & 0.881 & 0.875 & 0.899 & 0.953 \\
Reacher & 0.351 & 0.033 & 0.291 & 0.514 \\
OGB-Cube    & 0.268 & 0.323 & 0.312 & 0.561 \\
\bottomrule
\end{tabular}
\end{table}

The same factor-wise interpretation is weaker on Reacher and OGB-Cube. This does not
imply that the learned state is non-physical: complementary probes recover
finger position in Reacher with correlation $0.999$, and end-effector and block
position in OGB-Cube with correlations $0.997$ and $0.989$, respectively. Rather,
these tasks appear to represent the variables useful for planning more
holistically in $s$, without requiring each of the $H$, $R$, and $G$ components to
be linearly aligned with a privileged physical coordinate.

\myparagraph{Interpretation.}
These results suggest two levels at which the structured predictor is useful.
First, the port-Hamiltonian parameterization provides a calibrated decomposition
into passive, dissipative, and directly controlled change whose relative use
adapts across tasks. Second, when the underlying task admits a simple physical
factorization, as in PushT, the learned fields themselves become strongly
interpretable in terms of physical evolution. Successful planning does not
require this stronger property: on Reacher and OGB-Cube, control-relevant physical
information remains clearly present in the planning state even when it is
distributed across the nonlinear predictor. We therefore view the
port-Hamiltonian structure primarily as an inductive bias that organizes latent
dynamics, with physical interpretability emerging where supported by the task
rather than being imposed by construction.

\section{History mechanism ablation}
\label{app:history}

An earlier variant used a three-frame history fused by a learned mixer with LayerNorm, $c=\mathrm{LN}(W_{\mathrm{hist}}[s,\Delta_1,\Delta_2])$. After training, about $90$ percent of the mixer's Frobenius energy concentrated on the latest state, and zeroing the difference channels barely changed one-step predictions (cosine similarity about $0.99$), so the mixer was learning to ignore the history it was built to fuse. The current parameter-free phase $c=[s,v]$ with a two-frame window removes the mixer and its normalization entirely, keeps $s$ a single-frame representation, and leaves the planner's metric untouched, and it is the variant used in all selected recipes.

\section{Geometry of the inherited control state}
\label{app:geometry}

This appendix makes the inheritance statement of \S\ref{sec:bottleneck} exact and compares it with trained models.

\myparagraph{Target covariance of the state.}
Let $e=\one/\sqrt{D}$. $\LN_0$ gives $e^{\top}h=0$ for every sample, so $\Cov(h)\,e=0$ for any encoder, and the target of \eqref{eq:iso} is $P=I-ee^{\top}$. The exact covariance of a uniform distribution on the radius-$\sqrt{D}$ sphere inside the hyperplane is $\tfrac{D}{D-1}P$, which differs from $P$ by $0.5$ percent at $D=192$, and we use the unit target as a fixed convention. If $\Cov(h)=P$, then by \eqref{eq:inherit}
\begin{equation}
\Cov(s)=U^{\top}(I-ee^{\top})U=I_r-ww^{\top},\qquad w=U^{\top}e .
\end{equation}
This matrix has eigenvalue $1$ with multiplicity $r-1$ and one exceptional eigenvalue $1-\|w\|_2^2$ along $w$. For $U$ drawn uniformly among orthonormal frames, $\|w\|_2^2$ follows a $\mathrm{Beta}\bigl(\tfrac r2,\tfrac{D-r}2\bigr)$ distribution with mean $r/D$. When $r=D$, $U$ is an orthogonal matrix, $\|w\|_2=1$, and the exceptional eigenvalue is exactly zero: the state then has one null direction, the image of the centering direction.

\begin{table}[h]
\centering
\caption{Measured geometry of the control state on $n=512$ encoded frames from the selected checkpoints. The Marchenko-Pastur (MP) interval is the range of sample eigenvalues that an exactly isotropic state would show at this $n$ and $r$ \citep{marchenko1967}.}
\label{tab:geometry}
\small
\begin{tabular}{@{}lccccccc@{}}
\toprule
 & & & exceptional & \multicolumn{3}{c}{measured eigenvalues of $\Cov(s)$} & \\
\cmidrule(lr){5-7}
Model & $\|U^{\top}e\|_2^2$ & $r/D$ & $1-\|U^{\top}e\|_2^2$ & min & median & max & MP interval \\
\midrule
PushT, $r=64$ & $0.356$ & $0.333$ & $0.644$ & $0.236$ & $0.713$ & $2.48$ & $[0.42,\,1.83]$ \\
PushT, $r=192$ & $1.000$ & $1.000$ & $0.000$ & $1.3\times10^{-6}$ & $0.877$ & $3.31$ & $[0.15,\,2.60]$ \\
OGB-Cube, $r=32$ & $0.116$ & $0.167$ & $0.884$ & $0.015$ & $0.625$ & $1.46$ & $[0.56,\,1.56]$ \\
\bottomrule
\end{tabular}
\end{table}

\begin{figure}[h]
\centering
\setlength{\tabcolsep}{2pt}
\begin{tabular}{@{}ccc@{}}
\includegraphics[width=0.32\textwidth]{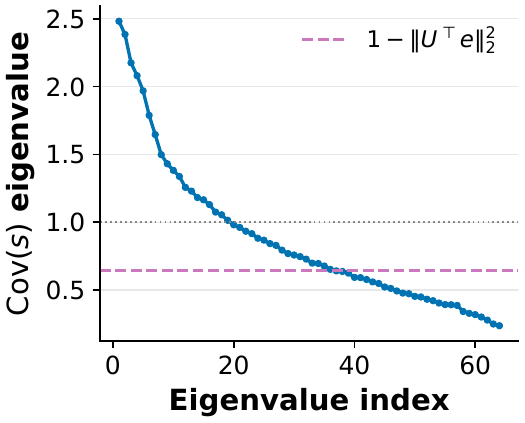} &
\includegraphics[width=0.32\textwidth]{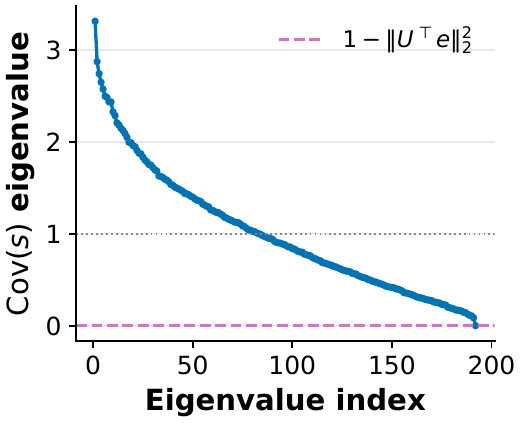} &
\includegraphics[width=0.32\textwidth]{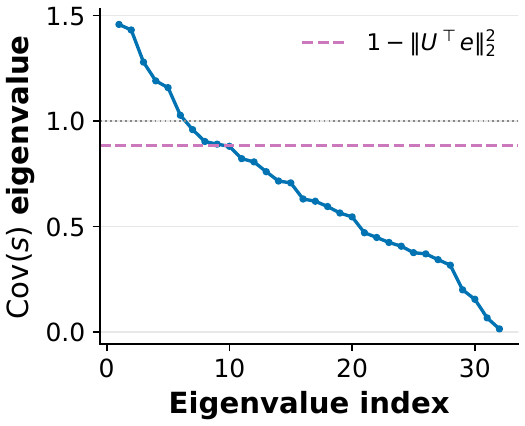} \\
\textbf{(a)} PushT, $r=64$ &
\textbf{(b)} PushT, $r=192$ &
\textbf{(c)} OGB-Cube, $r=32$
\end{tabular}
\caption{\textbf{Eigenvalues of $\Cov(s)$ for the selected checkpoints.} The dashed line marks the exceptional eigenvalue $1-\|U^{\top}e\|_2^2$ of the target covariance and the dotted line marks one. At $r=D=192$ the predicted null direction appears as the last eigenvalue.}
\label{fig:cov-s}
\end{figure}

\myparagraph{What trained models show.}
Table~\ref{tab:geometry} and Figure~\ref{fig:cov-s} support three statements. First, the measured $\|U^{\top}e\|_2^2$ is close to $r/D$ for each seeded projection, as expected. Second, the structural prediction at $r=D$ is confirmed: the smallest eigenvalue of $\Cov(s)$ is $1.3\times10^{-6}$, so the full-rank PushT state lives in a hyperplane of dimension $D-1$. Third, for $r<D$ every state direction retains nonzero variance, but the spectrum is spread around one and is not flat, and the smallest eigenvalue on OGB-Cube is well below the Marchenko-Pastur range. Part of this spread is a finite-sample effect, as the MP intervals show, and part is residual anisotropy of the code, whose own covariance eigenvalues reach about $2.75$ in these checkpoints. The smallest measured eigenvalue lies below the exceptional value of the target in both $r<D$ models, so the tail of the spectrum is set by this residual anisotropy and not by the centering direction. The tail is lowest on OGB-Cube, whose selected checkpoint is from epoch $3$. We therefore describe the state as inheriting a full-rank, well-scaled covariance from the code, which is exactly isotropic only at the prior's optimum.

\section{Anti-collapse penalties near vanishing variance}
\label{app:collapse}

\begin{figure}[ht]
\centering
\includegraphics[width=0.45\textwidth]{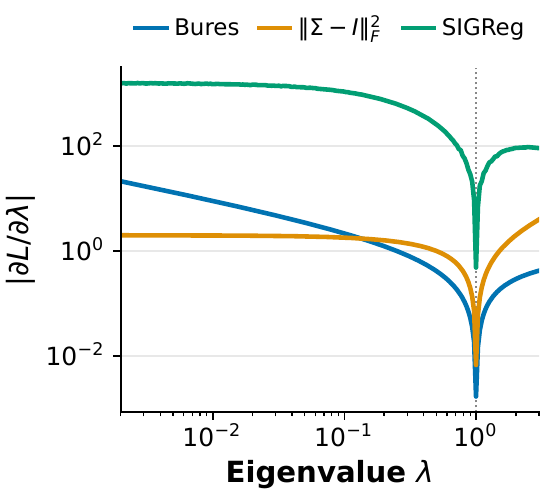}
\caption{\textbf{Eigenvalue derivatives of the three penalties on synthetic data.} The plot shows $|\partial\ell/\partial\lambda|$ against $\lambda$. Bures-Wasserstein rises as $\lambda^{-1/2}$ when $\lambda\to0$, while the Euclidean penalty and SIGReg flatten to constants. By \eqref{eq:chain}, a flat curve means a force on the standard deviation that vanishes like $\sqrt{\lambda}$, and the $\lambda^{-1/2}$ rise means a constant force. Vertical offsets reflect the native normalization of each statistic and are not comparable across curves. The comparison is between slopes.}
\label{fig:collapse}
\end{figure}

Consider one covariance direction with variance $\lambda$ and standard deviation $\sigma=\sqrt{\lambda}$, and write the samples along it as $z_j=\sigma x_j$ with standardized $x_j$. For any penalty $\ell$, the chain rule gives
\begin{equation}
\frac{\partial \ell}{\partial\sigma}=2\sigma\,\frac{\partial \ell}{\partial\lambda},
\label{eq:chain}
\end{equation}
so a penalty whose eigenvalue derivative stays bounded as $\lambda\to0$ exerts a force on $\sigma$ that vanishes linearly, and only a derivative that diverges like $\lambda^{-1/2}$ keeps a constant force. The three penalties of \eqref{eq:forces} fall on the two sides of this line, as Figure~\ref{fig:collapse} shows.

\myparagraph{Bures-Wasserstein.} $\ell=(\sqrt{\lambda}-1)^2$ has $\partial_\lambda\ell=1-\lambda^{-1/2}$, which diverges at exactly the compensating rate, and $\partial_\sigma\ell=2(\sigma-1)\to-2$.

\myparagraph{Euclidean covariance penalty.} $\ell=(\lambda-1)^2$ has $\partial_\lambda\ell=2(\lambda-1)\to-2$, bounded, and $\partial_\sigma\ell=4\sigma(\sigma^2-1)\to0$.

\myparagraph{SIGReg.} The Epps-Pulley statistic \citep{epps1983test} compares the empirical characteristic function $\varphi_n(\tau)=\tfrac1n\sum_j \exp(\mathrm{i}\tau\sigma x_j)$ with that of a standard Gaussian under a weight $\omega$, $T(\sigma)=n\int|\varphi_n(\tau)-e^{-\tau^2/2}|^2\,\omega(\tau)\,d\tau$. Expanding at small $\sigma$ with standardized $x_j$ gives $\varphi_n(\tau)=1-\tfrac12\tau^2\sigma^2+O(\sigma^3)$, where the cubic term is purely imaginary, hence
\begin{equation}
T(\sigma)=T(0)-\kappa\,\sigma^2+O(\sigma^4),\qquad
\kappa=n\int \tau^2\bigl(1-e^{-\tau^2/2}\bigr)\omega(\tau)\,d\tau>0 .
\end{equation}
The statistic is bounded, since $|\varphi_n(\tau)-e^{-\tau^2/2}|\le2$, its eigenvalue derivative tends to the constant $-\kappa$, and its force on $\sigma$ is $-2\kappa\sigma\to0$. A collapsing direction therefore sees a saturated SIGReg score and a vanishing restoring force, as we observed in training.

\section{Training behavior}
\label{app:training}

\begin{figure*}[t]
\centering
\setlength{\tabcolsep}{2pt}
\begin{tabular}{@{}ccc@{}}
\includegraphics[width=0.31\textwidth]{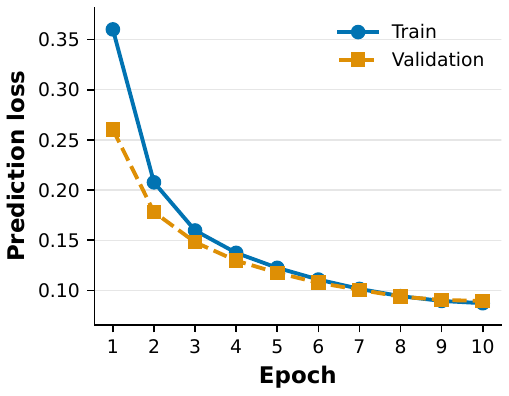} &
\includegraphics[width=0.31\textwidth]{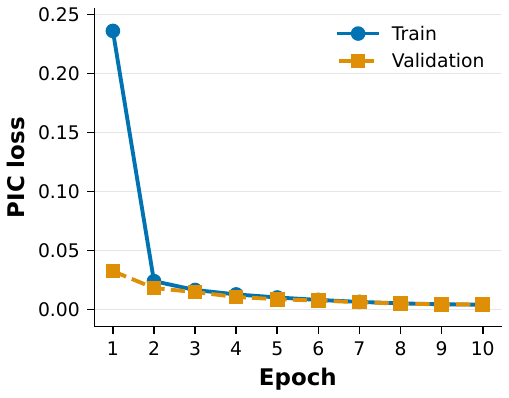} &
\includegraphics[width=0.31\textwidth]{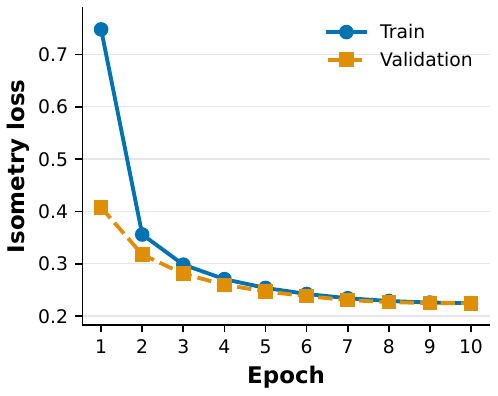} \\
\textbf{(a)} Prediction &
\textbf{(b)} Port-inverse consistency &
\textbf{(c)} Isotropy
\end{tabular}
\caption{\textbf{H-JEPA training on PushT.}
Training and validation losses across the $10$-epoch run for
\textbf{(a)} open-loop latent prediction,
\textbf{(b)} port-inverse consistency (PIC), and
\textbf{(c)} the Bures-Wasserstein isotropy objective.
All objectives decrease smoothly, with closely tracking training and validation
curves.}
\label{fig:train-pusht}
\end{figure*}

\begin{figure*}[t]
\centering
\setlength{\tabcolsep}{2pt}
\begin{tabular}{@{}ccc@{}}
\includegraphics[width=0.31\textwidth]{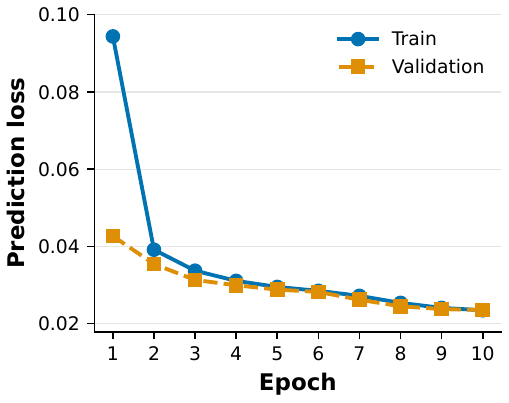} &
\includegraphics[width=0.31\textwidth]{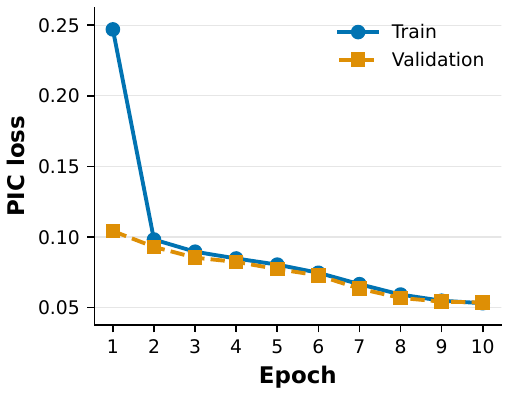} &
\includegraphics[width=0.31\textwidth]{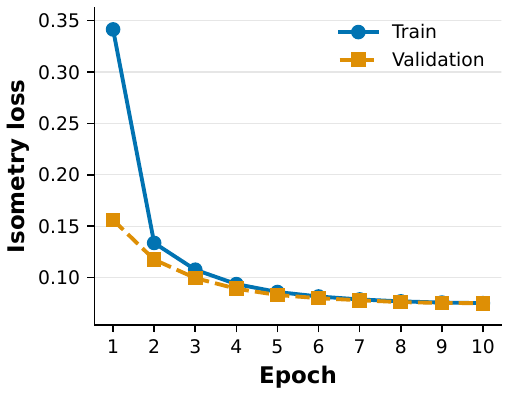} \\
\textbf{(a)} Prediction &
\textbf{(b)} Port-inverse consistency &
\textbf{(c)} Isotropy
\end{tabular}
\caption{\textbf{H-JEPA training on OGB-Cube.}
Training and validation losses for
\textbf{(a)} open-loop latent prediction,
\textbf{(b)} PIC, and
\textbf{(c)} isotropy.
All three objectives fall rapidly during the first epochs and subsequently
improve smoothly, with little separation between training and validation curves.
The benchmark model uses the epoch-$3$ OGB-Cube checkpoint selected
in \S\ref{sec:setup}. The extended trace shows that lower representation-level
losses after convergence are not required for the reported planning result.}
\label{fig:train-cube}
\end{figure*}

All H-JEPA models were trained end-to-end for $10$ epochs on a single NVIDIA RTX A5000 GPU with the optimization settings of Appendix~\ref{app:config}. A held-out pilot seed, excluded from the three-seed benchmark average, was used once to choose which epoch's checkpoint to evaluate on each task. We then fixed this choice at epoch $8$ for Two-Room, $6$ for Reacher, $10$ for PushT, and $3$ for OGB-Cube for every reported seed. For comparison, Delta-JEPA is trained for $50$ epochs under its evaluation protocol \citep{deltajepa2026}. Figures~\ref{fig:train-pusht} and~\ref{fig:train-cube} show the available extended traces for PushT and OGB-Cube, including prediction, PIC, and Bures-Wasserstein isotropy on both training and validation data.

The curves remain closely matched between training and validation. PushT improves steadily over all $10$ epochs. OGB-Cube reaches its useful regime within the first three epochs, after which the representation-level losses continue to decrease without a corresponding need for a later checkpoint. The traces also show that all three objectives can be optimized jointly without loss-specific warm-up schedules or alternating optimization.

\end{document}